\documentclass[letterpaper]{article}

\usepackage{aaai18}
\usepackage{times}
\usepackage{helvet}
\usepackage{courier}
\usepackage{url}
\usepackage{graphicx}
\usepackage{mathtools}
\usepackage{amsmath,amssymb,amsthm,xfrac,amsbsy}
\usepackage[tight]{subfigure}
\usepackage[ruled]{algorithm2e}
\usepackage{dsfont}
\usepackage{tabulary}

\title{Learning User Preferences to Incentivize Exploration \\ in the Sharing Economy}
\author{
Christoph Hirnschall \\ ETH Zurich \\ Zurich, Switzerland \\ chirnsch@gmail.com
\And 
Adish Singla \\ MPI-SWS \\ Saarbr{\"u}cken, Germany \\ adishs@mpi-sws.org
\And
Sebastian Tschiatschek \\ Microsoft Research\thanks{Work performed while at ETH Zurich.} \\ Cambridge, United Kingdom \\ setschia@microsoft.com
\And
Andreas Krause \\ ETH Zurich \\ Zurich, Switzerland \\ krausea@ethz.ch}

\newcommand{\problems}{K}
\newcommand{\solutionspace}{S}
\newcommand{\loss}{l}
\newcommand{\prediction}{p}
\newcommand{\w}{\pmb{w}}
\newcommand{\wtilde}{\widetilde{\w}}
\newcommand{\wstar}{\pmb{u}}
\newcommand{\Q}{\pmb{Q}}
\newcommand{\gradient}{\pmb{g}}
\newcommand{\z}{z}
\newcommand{\counter}{\tau}
\newcommand{\maxgradient}{\gradient}
\newcommand{\maxdualitygap}{\delta^t}
\newcommand{\dualitygap}{\delta}
\newcommand{\calpha}{c_\alpha}
\newcommand{\cbeta}{c_\beta}

\newcommand{\OCP}{OL\xspace}
\newcommand{\IOCP}{IOL\xspace}
\newcommand{\COCP}{CoOL\xspace}

\newcommand{\aprxproj}{AProj\xspace}

\newcommand{\bad}{\mathds{1}_{\{(\neg \xi^{t-1}) \wedge (\xi^{t})\}}}
\newcommand{\good}{\mathds{1}_{\{(\xi^{t-1}) \vee (\neg \xi^{t})\}}}
\newcommand{\outside}{\mathds{1}_{\{\neg \xi^t\}}}
\newcommand{\inside}{\mathds{1}_{\{\xi^t\}}}
\newcommand{\insidet}[1]{\mathds{1}_{\{\xi^{#1}\}}}
\newcommand{\outsidet}[1]{\mathds{1}_{\{\neg \xi^{#1}\}}}

\newcommand{\argmin}{\mathop{\mathrm{argmin}}\limits} 
\newcommand{\norm}[1]{\left\lVert#1\right\rVert}
\newcommand{\names}[1]{#1}

\newcommand{\regret}{Regret}
\newcommand{\expectedvalue}[1]{\mathbf{E} \left[#1\right]}

\DeclarePairedDelimiterX{\normnew}[1]{\big\lVert}{\big\rVert}{#1}

\newcommand{\what}{\widehat{\w}}
\newcommand{\x}{\pmb{x}}
\newcommand{\zt}{{z^t}}

\newcommand{\cost}{{c}}

\newcommand{\utility}{u}

\newcommand{\regularizer}{R}
\newcommand{\divergence}{D}
\newtheorem{theorem}{Theorem\xspace}
\newtheorem{proposition}{Proposition\xspace}
\newtheorem{lemma}{Lemma\xspace}
\newtheorem{corollary}{Corollary\xspace}

\begin{document}

\maketitle

%%%%%%%%%%%%%%%%%%%%%%%%%%%%%%%%%%%%%%%%%%%%%%%%%%%%%%%%%%%
%%%%%%%%%%%%%%%%%%%%%%%%%%%%%%%%%%%%%%%%%%%%%%%%%%%%%%%%%%% ABSTRACT
\begin{abstract}
We study platforms in the sharing economy and discuss the need for incentivizing users to explore options that otherwise would not be chosen. For instance, rental platforms such as Airbnb typically rely on customer reviews to provide users with relevant information about different options. Yet, often a large fraction of options does not have any reviews available. Such options are frequently neglected as viable choices, and in turn are unlikely to be evaluated, creating a vicious cycle. Platforms can engage users to deviate from their preferred choice by offering monetary incentives for choosing a different option instead. To efficiently learn the optimal incentives to offer, we consider structural information in user preferences and introduce a novel algorithm - Coordinated Online Learning (CoOL) - for learning with structural information modeled as convex constraints. We provide formal guarantees on the performance of our algorithm and test the viability of our approach in a user study with data of apartments on Airbnb. Our findings suggest that our approach is well-suited to learn appropriate incentives and increase exploration on the investigated platform.
\end{abstract}

%%%%%%%%%%%%%%%%%%%%%%%%%%%%%%%%%%%%%%%%%%%%%%%%%%%%%%%%%%%
%%%%%%%%%%%%%%%%%%%%%%%%%%%%%%%%%%%%%%%%%%%%%%%%%%%%%%%%%%% INTRODUCTION
\section{Introduction}\label{sec.introduction}
In recent years, numerous sharing economy platforms with a variety of goods and services have emerged. These platforms are shaped by users that primarily act in their own interest to maximize their utility. However, such behavior might interfere with the usefulness of the platforms. For example, users of mobility sharing systems typically prefer to drop off rentals at the location in closest proximity, while a more balanced distribution would allow the mobility sharing service to operate more efficiently.

Undesirable user behavior in the sharing economy is in many cases even self-reinforcing. For example, users in the apartment rental marketplace Airbnb are less likely to select infrequently reviewed apartments and are therefore unlikely to provide reviews for these apartments
\cite{fradkin2014search}. This is also reflected in the distribution of reviews, where in many cities $20\%$ of apartments account for more than $80\%$ of customer reviews\footnote{Data from \url{insideairbnb.com}.}.

Such dynamics create a need for platforms in the sharing economy to actively engage users to shape demand and improve efficiency. Several previous papers have proposed the idea of using monetary incentives to encourage desirable behavior in such systems. One example is \cite{frazier2014incentivizing}, who studied the problem in a multi-armed bandit setting, where a principal (\emph{e.g.} a marketplace) attempts to maximize utility by incentivizing agents to explore arms other than the myopically preferred one. In their setting, the optimal amount is known to the system, and the main goal is to quantify the required payments to achieve an optimal policy with myopic agents. The idea of shaping demand through monetary incentives in the sharing economy has also been tested in practice. For example, \cite{singla2015incentivizing} use monetary incentives to encourage users of bike sharing systems to return bikes at beneficial locations, making automatic offers through the bike sharing app.

In this context, an important question is what amounts a platform should offer to maximize its utility. \cite{singla2015incentivizing} introduce a simple protocol for learning optimal incentives in the bike sharing system to make users switch from the preferred station to a more beneficial one, ignoring information about specific switches and additional context.
Extending on these ideas, we explore a general online learning protocol for efficiently learning optimal incentives.

\subsection{Our Contributions}
We provide the following main contributions in this paper:

\begin{itemize}
  \item {\bfseries Structural information:} We consider structural information in user preferences to speed up learning of incentives, and provide a general framework to model structure across tasks via convex constraints. Our algorithm, Coordinated Online Learning (\COCP) is also of interest for related multi-task learning problems.
  \item {\bfseries Computational efficiency:} We introduce two novel ideas of \emph{sporadic} and \emph{approximate} projections to increase the computational efficiency of our algorithm. We derive formal guarantees on the performance of the \COCP algorithm and achieve no-regret bounds in this setting.
  \item {\bfseries User study on Airbnb:} We collect a unique data set through a user study with apartments on Airbnb and test the viability and benefit of the \COCP algorithm on this dataset.
\end{itemize}

%%%%%%%%%%%%%%%%%%%%%%%%%%%%%%%%%%%%%%%%%%%%%%%%%%%%%%%%%%%%
%%%%%%%%%%%%%%%%%%%%%%%%%%%%%%%%%%%%%%%%%%%%%%%%%%%%%%%%%%%% Preliminaries
\section{Preliminaries}\label{sec.model}
In the following, we introduce the general problem setting of this paper.

{\bfseries Platform.} We investigate a general platform in the sharing economy, such as the apartment rental marketplace \names{Airbnb}. On this platform, users can choose from $n$ goods and services, denoted as items. A user that arrives at time $t$ chooses an item $i^t \in [n]$. If the user chooses to buy item $i^t$, the platform gains utility $u^t_i$.

{\bfseries Incentivizing exploration.} The initial choice, item $i^t$, might not maximize the platform's utility, and the platform might be interested in offering a different item $j^t$ with utility $u^t_j > u^t_i$ instead. For example, $j$ could represent an infrequently reviewed item that the platform wants to explore. To motivate the user to select item $j^t$ instead, the platform can offer an incentive $\prediction^t$, for example in the form of a monetary discount on that item. The user can either accept or reject the offer $\prediction^t$ depending on the private cost $\cost^t$, where the user accepts the offer if $\prediction^t \geq \cost^t$ and rejects the offer otherwise. If the user accepts the offer, the utility gain of the platform is $u^t_j - u^t_i - \prediction^t$. %This setting is illustrated in Figure~\ref{fig.incentives}.

{\bfseries Objective.} In this setting, two tasks need to be optimized to achieve a high utility gain: finding good switches $i \rightarrow j$, and finding good incentives $\prediction^t$. Good switches $i \rightarrow j$ are those, in which the achievable utility gain is positive, i.e. $u^t_j - u^t_i - \cost^t > 0$. To realize a positive utility gain, the offer $\prediction^t$ needs to be greater or equal to $\cost^t$, since otherwise the offer would be rejected.

In this paper, we focus on learning optimal incentives $\prediction^t$ over time, while the platform chooses relevant switches $i \rightarrow j$ independently.

%%%%%%%%%%%%%%%%%%%%%%%%%%%%%%%%%%%%%%%%%%%%%%%%%%%%%%%%%%%%
%%%%%%%%%%%%%%%%%%%%%%%%%%%%%%%%%%%%%%%%%%%%%%%%%%%%%%%%%%%% Methodology
\section{Methodology}\label{sec.algorithm}
In this section, we present our methology for learning optimal incentives $\prediction^t_{i,j}$ and start with a single pair of items $(i,j)$. We allow for natural constraints on $\prediction^t_{i,j}$, such that $\prediction^t_{i,j} \in \solutionspace_{i,j}$, where $\solutionspace_{i,j}$ is convex and non-empty. For example, $\solutionspace_{i,j}$ might be lower-bounded by $0$ and upper-bounded by the maximum discount that the platform is willing to offer.

\subsection{Single Pair of Items}
We consider the popular algorithmic framework of online convex programming (OCP) \cite{zinkevich2003online} to learn optimal incentives $\prediction^t_{i,j}$ for a single pair of items. The OCP algorithm is a gradient-descent style algorithm that updates with an adaptive learning rate and performs a projection after every gradient step to maintain feasibility within the constraints $\solutionspace_{i,j}$. We use $\counter^t_{i,j}$ to denote the number of times a pair of items ${i,j}$ has been observed and $\eta$ to denote the learning rate. To measure the performance of the algorithm, we use the loss $\loss^t(\prediction^t_{i,j})$, which is the difference between the optimal prediction and the prediction provided by the algorithm, such that $l^t(p^t) = \mathds{1}_{\{p^t \geq c^t\}} \cdot (p^t - c^t) + \mathds{1}_{\{p^t < c^t\}} \cdot (u - c^t)  \textnormal{ for } u \geq c^t$, and $l^t(p^t) = 0 \textnormal{ for } u < c^t$\footnote{Note that this loss function is non-convex. A convex version is presented in the case study.}. 

%%%%%%%%%%%%%%%%%%%%%%%%%%%%%%%%%%%%%%%%%%%%%%%%%%%%%%%%%%
\begin{algorithm}[t!]
\nl    {\bfseries Input:}  
    {    
    \begin{itemize}
             \item Learning rate constant: $\eta > 0$ 
    \end{itemize}
    }
\nl    {\bfseries Initialize:}  {$\prediction^0_{i,j} \in \solutionspace$, $\counter^0_{i,j} = 0$} \\
\nl    \For{$t = 1, 2, \ldots, T$}{ 
\nl            Suffer loss $\loss^t(\prediction^t_{i,j})$\\
\nl            Calculate gradient $g^t_{i,j}$\\
\nl            Set $\counter^t_{i,j} = \counter^{t-1}_{i,j} + 1$ \\
\nl            Update $\prediction^{t+1}_{i,j} = \prediction^t_{i,j} - \frac{\eta}{\sqrt{\counter^t_{i,j}}} g^t_{i,j}$
    }
\caption{\OCP \ -- Online Learning}
\label{algo.OCP}
\end{algorithm}
%%%%%%%%%%%%%%%%%%%%%%%%%%%%%%%%%%%%%%%%%%%%%%%%%%%%%%%%%%

The algorithm, Online Learning (\OCP), for a single pair of items is shown in Algorithm~\ref{algo.OCP}. The regret, which measures the difference in losses against any (constant) competing weight vector $\prediction_{i,j} \in  \solutionspace$ after $T$ rounds, is defined as
\begin{align}
\regret_{\OCP_{i,j}}(T, \prediction_{i,j}) = \sum_{t=1}^{T} \Big(\loss^t(\prediction^t_{i,j}) - \loss^t(\prediction_{i,j})\Big). \label{eq.regretOCP}
\end{align}

\subsection{Multiple Pairs of Items}
We now relax the assumption of a fixed pair of items and return to our original problem of learning optimal incentives for multiple pairs of items, \emph{i.e.} the algorithm receives specific items $i^t$ and $j^t$ as input for each user. If we consider all items $n$ on the platform, the total number of pairs is $n^2 - n$.

For learning the optimal incentive for each pair of items, the algorithm maintains a specific learning rate proportional to $\sfrac{1}{\sqrt{\counter^t_{i,j}}}$ for each pair of items and performs one gradient update step using Algorithm~\ref{algo.OCP}. We refer to this straightforward adaptation of the \OCP algorithm as Independent Online Learning (\IOCP) and use this algorithm as a baseline for our analysis. Using regret bounds of \cite{zinkevich2003online} and denoting the number of pairs of items as $\problems$, we can upper bound the regret of \IOCP as

\begin{align}
\regret_{\IOCP}(T) \leq \frac{3}{2} \sqrt{T \problems} \norm{\solutionspace_{max}} \norm{\maxgradient_{max}}.\end{align} 

\subsection{Structural Information} \label{sec.structureintro}
In a real-world setting, incentives for different pairs of items typically are not independent, and in some cases, certain structural information may help to speed up learning of optimal incentives. In the following, we discuss several relevant types of structural information.

{\bfseries Independent learning.} In this baseline setting each pair of items is learned individually. Thus, the number of incentives that need to be learned grows quadratically with the number of items on a platform. While applicable for a small number of items, this approach is not favorable on typical platforms in the sharing economy.
 
{\bfseries Shared learning.} Another commonly studied setting is shared learning. In this setting, all pairs of items are considered equivalent, and only one global incentive is learned. While allowing the platform to learn about many pairs of items at the same time, this approach fails to consider natural asymmetries in the problem. For example, the required incentive for switching from $i$ to $j$ is often different than the required incentive for switching from $j$ to $i$, as can be also observed in the case study of this paper.

{\bfseries Metric/hemimetric structure.} Assuming that the required incentives are related to the dissimilarity of items $i$ and $j$, metrics are a natural choice to model structural dependencies, as they capture the property of triangle inequalities in dissimilarity functions. However, incentives for pairs of items are not necessarily required to be symmetric. For example, the required incentives for switching from a highly reviewed apartment on Airbnb to one without reviews is likely higher than vice versa. Therefore, we use hemimetrics, which are a relaxed form of a metric that satisfy only non-negativity constraints and triangular inequalities, capturing asymmetries in preferences (\emph{cf.} \cite{singla2016actively}). The usefulness of the hemimetric structure for learning optimal incentives is demonstrated in the experiments.

In the following section, we introduce a general-purpose algorithm for learning with structural information, where the structure is defined by convex constraints on the solution space. The key idea of our algorithm is to coordinate between individual pairs of items by projecting onto the resulting convex set. We generalize our approach for contextual learning, where additional features, such as information about users, may be available. Since projecting onto convex sets may be computationally expensive, we further extend our analysis to allow projections to be sporadic (i.e. only after certain gradient steps) and approximate (i.e. with some error compared to the optimal projection).

%%%%%%%%%%%%%%%%%%%%%%%%%%%%%%%%%%%%%%%%%%%%%%%%%%%%%%%%%%%%
%%%%%%%%%%%%%%%%%%%%%%%%%%%%%%%%%%%%%%%%%%%%%%%%%%%%%%%%%%%% STRUCTURAL INFORMATION
\section{Learning with Structural Information}
We begin this section by introducing a general framework for specifying structural information via convex constraints.
We denote each pair of items as a distinct problem $\z \in [K]$, where $K$ is the total number of pairs of items. Each problem $\z$ may be associated with additional features, for example with information about the current user. As is common in online learning, we consider a $d$ dimensional weight vector $\w_\z$ for each problem $\z \in [\problems]$ for learning optimal incentives. The prediction $\prediction_\z$ is equal to the inner product between $\w_\z$ and the $d$ dimensional feature vector. In the previous section, we described the special case with $d=1$ and a unit feature vector, such that $\w_\z$ is equivalent to the prediction $\prediction_\z$.

\subsection{Specifying Structure via Convex Constraints}
Similar to constraints on $\prediction_\z$, we allow for convex constraints on $\w_\z$, such that $\w_\z \in \solutionspace_\z \subseteq \mathbb{R}^{d}$. We assume $\solutionspace_\z$ is a convex, non-empty, and compact set, where $\norm{\solutionspace_\z}$ is the Euclidean norm of the solution space\footnote{Euclidean norm is used throughout, unless otherwise specified.}. Further, we assume $\norm{\solutionspace_\z} \leq \norm{\solutionspace_{\textnormal{max}}}$ for some constant $\norm{\solutionspace_{\textnormal{max}}}$.
We denote the joint solution space of the $\problems$ problems as $\solutionspace =  \solutionspace_1 \times {\cdots} \times \solutionspace_\z  \times {\cdots} \times \solutionspace_\problems \subseteq \mathbb{R}^{d \cdot {\problems}}$ and define $\w^t \in \solutionspace$ as the concatenation of the problem specific weight vectors at time $t$, \emph{i.e.} 
\begin{align*}
\w^t=\big[(\w^t_1)' \ {\cdots} (\w^t_\z)' \ {\cdots} (\w^t_\problems)'\big]'.
\end{align*}

The available structural information is modelled by a set of convex constraints, such that the joint competing weight vector $\w^*$, against which the loss at each round is measured, lies in a convex, non-empty, and closed set $\solutionspace^* \subseteq \solutionspace$, representing a restricted joint solution space, \emph{i.e.} $\w^* \in \solutionspace^*$. In the following, we provide several practical examples of how $\solutionspace^*$ can be defined.

{\bfseries Independent learning.} $\solutionspace^* \equiv \solutionspace$ models the setting where the problems are unrelated/independent.

{\bfseries Shared learning.} A shared parameter setting can be modeled as
\begin{align*}
\solutionspace^* = \{\w^* \in \solutionspace \ | \ \w^*_1 = \cdots = \w^*_\z  = \cdots = \w^*_\problems\}.
\end{align*}
Instead of sharing all parameters, another common scenario is to share only a few parameters. For a given $d' \leq d$, sharing $d'$ parameters across the problems can be modeled as 
\begin{align*}
\solutionspace^* = \{\w^* \in \solutionspace \ | \ \w^*_1[1\!:\!d']  \cdots = \w^*_\z[1\!:\!d']  = \cdots \w^*_\problems[1\!:\!d']\}
\end{align*}
where $\w^*_\z[1\!:\!d']$ denotes the first $d'$ entries in $\w^*_\z$. This approach is useful for sharing certain parameters that do not depend on the specific problem. For example, in the case of apartments on \names{Airbnb}, a shared feature could be the distance between apartments.

{\bfseries Hemimetric structure.} To model dissimilarities between items for learning optimal incentives, we use the hemimetric set. Specifically, we use r-bounded hemimetrics, which, next to non-negativity constraints and triangular inequalities, also include non-negativity and upper bound constraints.
For $d=1$, the convex set representing $r$-bounded hemimetrics is given by $\solutionspace^* =$
\begin{align*}
\{\w^* \in \solutionspace \ | \ \w^*_{i,j} \in [0,r], \w^*_{i,j} \leq \w^*_{i,k} + \w^*_{k,j} \ \forall i, j, k \in [n]\}
\end{align*}

\subsection{Our Algorithm}
In the following, we introduce our algorithm, Coordinated Online Learning (\COCP). 

{\bfseries Exploiting Structure via Weighted Projections.}
The \COCP algorithm exploits structural information in a principled way by performing weighted projections to $\solutionspace^*$, with weights for a problem $\z$ proportional to $\sqrt{\counter^t_\z}$. Intuitively, the weights allow us to learn about problems that have been observed infrequently while avoiding to ``unlearn" problems that have been observed more frequently. A formal justification for using weighted projections is provided in the extended version of this paper \cite{hirnschall2017learning}. 

We define $Q^t$ as a square diagonal matrix of size $d \problems$ with each $\sqrt{\counter^t_\z}$ represented $d$ times. In the one-dimensional case ($d=1$), we can write $Q^t$ as
\begin{align}
\Q^{t} = 
\begin{bmatrix}
\sqrt{\counter^t_1} & & 0  \\
& \ddots \\
0 & & \sqrt{\counter^t_\problems}
\end{bmatrix}.
\label{eq.qmatrix}
\end{align}
Using $\wtilde$ to jointly represent the current  weight vectors of all the learners at time $t$ (\emph{cf.}\ Line~\ref{alg1.update} in Algorithm~\ref{algo.COCP}), we compute the new joint weight vector $\w^{t+1}$ (\emph{cf.}\ Line~\ref{alg1.wnew} in Algorithm~\ref{algo.COCP}) by projecting onto $\solutionspace^*$, using
\begin{align}
\w^{t+1} &= \argmin_{\w \in \solutionspace^*} (\w - \wtilde)' \Q^t (\w - \wtilde). \label{eq.weightedproj}
\end{align}

We refer to this as the weighted projection onto $\solutionspace^*$. Since $\solutionspace^*$ is convex and the projection is a special case of the Bregman projection,
the projection onto $\solutionspace^*$ is unique (\emph{cf.} \cite{cesa2006prediction,rakhlin2009lecture}).

{\bfseries Sporadic and Approximate Projections.}
For large scale applications (\emph{i.e.} large $\problems$ or large $d$), projecting at every step could be computationally very expensive: a projection onto a generic convex set $\solutionspace^*$ would require solving a quadratic program of dimension $d \problems$. To allow for computationally efficient updates, we introduce two novel algorithmic ideas: \emph{sporadic} and \emph{approximate} projections, defined by the above-mentioned sequences $(\xi^t)_{t \in [T]}$  and $(\dualitygap^t)_{t \in [T]}$. 
Here, $\dualitygap^t$ denotes the desired accuracy at time $t$ and is given as input to Function~\ref{algo.aprxproj}, \aprxproj, for computing approximate projections. This way, the accuracy can be efficiently controlled using the duality gap of the projections.  As we shall see in our experimental results, these two algorithmic ideas of sporadic and approximate projections allow us to speed up the algorithm by an order of magnitude while retaining the improvements obtained through the projections.

Algorithm~\ref{algo.COCP}, when invoked with $\xi^t = 1, \dualitygap^t = 0  \ \forall t \in [T]$, corresponds to a variant of our algorithm with exact projections at every time step. When invoked with $\xi^t = 0 \ \forall t \in [T]$, our algorithm corresponds to the \IOCP baseline.

%%%%%%%%%%%%%%%%%%%%%%%%%%%%%%%%%%%%%%%%%%%%%%%%%%%%%%%%%%
\begin{algorithm}[t!]
\nl    {\bfseries Input:}  
    {    
    \begin{itemize}
            \item Projection steps: $(\xi^t)_{t \in [T]} \textnormal{ where } \xi^t \in \{0,1\}$ \\
            \item Projection accuracy:  $(\dualitygap^t)_{t \in [T]} \textnormal{ where } \dualitygap^t \geq 0$ \\
             \item Learning rate constant: $\eta > 0$ 
    \end{itemize}
    }
\nl    {\bfseries Initialize:}  {$\w^1_\z \in \solutionspace_\z$, $\counter^0_\z = 0$} \\
\nl    \For{$t = 1, 2, \ldots, T$}{ 
\nl            Suffer loss $\loss^t(\w^t_{\z})$\\
\nl            Calculate (sub-)gradient $\gradient^t_{\z}$\\
\nl            Set $\counter^t_{\z} = \counter^{t-1}_{\z} + 1$ \\
\nl            Update $\wtilde^{t+1} = \w^t$; $\wtilde^{t+1}_{\z} = \w^t_{\z} - \frac{\eta}{\sqrt{\counter^t_{\z}}} \gradient^t_{\z}$  \label{alg1.update}\\ 
\nl            \eIf{$\xi^t = 1$}{
\nl            Define $\Q^t$ as per Equation~\eqref{eq.qmatrix} \label{alg1.Q} \\
\nl            Compute $\w^{t+1} = \aprxproj(\wtilde^{t+1}, \dualitygap^t, \Q^t)$  \label{alg1.wnew}  \\
            }
            {
\nl                $\w^{t+1}_\z = \argmin_{\w \in \solutionspace_\z}  \normnew{\w - \wtilde^{t+1}_\z}_2$    \\
    }
    }
\caption{\COCP \ -- Coordinated Online Learning}
\label{algo.COCP}
\end{algorithm}
%%%%%%%%%%%%%%%%%%%%%%%%%%%%%%%%%%%%%%%%%%%%%%%%%%%%%%%%%%

%%%%%%%%%%%%%%%%%%%%%%%%%%%%%%%%%%%%%%%%%%%%%%%%%%%%%%%%%%
{\renewcommand{\algorithmcfname}{Function}
\begin{algorithm}[t!]

\nl     {\bfseries Input:} $\widetilde{\w}, \dualitygap^t, \Q^t$\\
\nl    Define $f^t(\w) = (\w - \wtilde)' \Q^t (\w - \wtilde)$ for $\w \in \solutionspace$\\
\nl    Choose $\w^{t+1} \in \{\w \in \solutionspace^* : f^t(\w) - \min_{\w' \in \solutionspace^*}\limits  f^t(\w') \leq \maxdualitygap \}$\\
\nl {\bfseries Return:} $\w^{t+1}$\\
\caption{\aprxproj \ -- Approximate Projection} 
\label{algo.aprxproj}
\end{algorithm}}
%%%%%%%%%%%%%%%%%%%%%%%%%%%%%%%%%%%%%%%%%%%%%%%%%%%%%%%%%%

{\bfseries Relation to existing approaches.} A related algorithm is the \names{AdaGrad} algorithm \cite{duchi2011adaptive}, which uses the sum of the magnitudes of past gradients to determine the learning rate at each time $t$, where larger past gradients correspond to smaller learning rates. A key difference to the \COCP algorithm is that the \names{AdaGrad} algorithm enforces exact projections after every iteration. This is particularly problematic for large, complex structures since projections on these structures often rely on numeric approximations, that may not guarantee to converge to the exact solution in finite time.

%%%%%%%%%%%%%%%%%%%%%%%%%%%%%%%%%%%%%%%%%%%%%%%%%%%%%%%%%%%%
%%%%%%%%%%%%%%%%%%%%%%%%%%%%%%%%%%%%%%%%%%%%%%%%%%%%%%%%%%%% ALGORITHM GUARANTEES

\section{Performance Guarantees and Analysis}\label{sec.analysis}
\begin{figure*}[!t] 
\centering
   \subfigure[Random problem order]{
     \includegraphics[trim = 2.5mm 3.2mm 4mm 10.5mm, clip=true, width=0.32\textwidth]{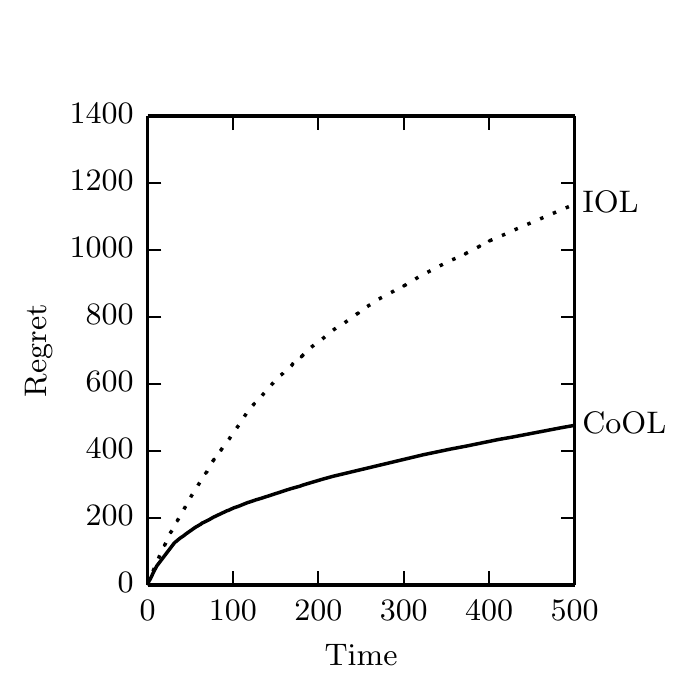}
    \label{fig.random}
   }   
   \subfigure[Batches of problems]{
     \includegraphics[trim = 2.5mm 3.2mm 4mm 10.5mm, clip=true, width=0.32\textwidth]{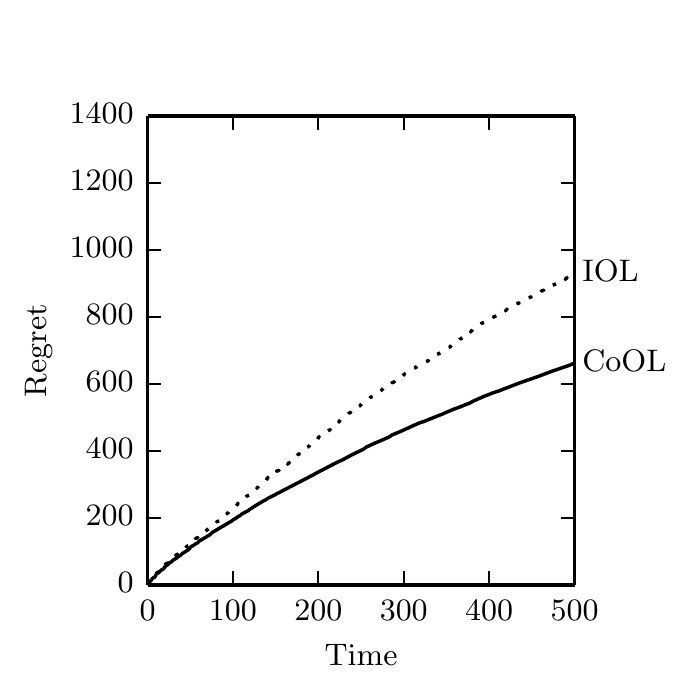}
     \label{fig.batch}
   } 
   \subfigure[Single problem]{
     \includegraphics[trim = 2.5mm 3.2mm 4mm 10.5mm, clip=true, width=0.32\textwidth]{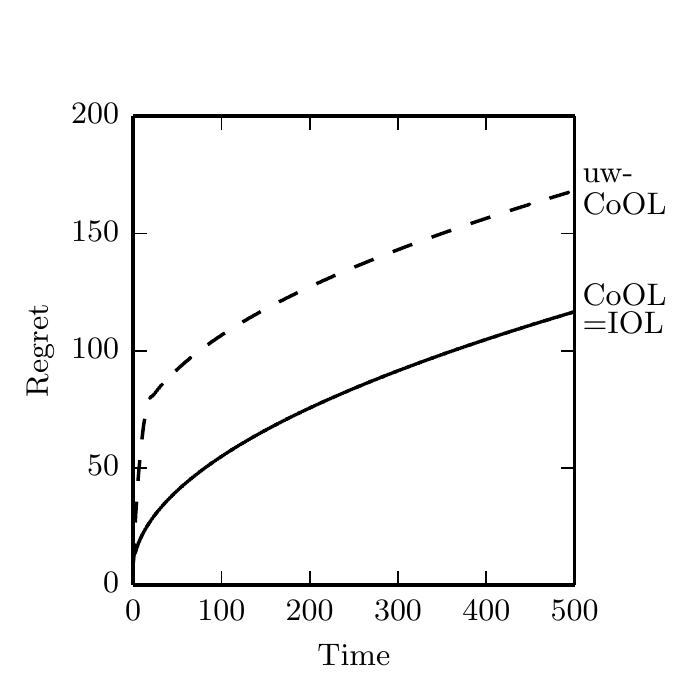}
     \label{fig.single}
   }
   \subfigure[Varying $\alpha$]{
     \includegraphics[trim = 2.5mm 3.2mm 4mm 10.5mm, clip=true, width=0.32\textwidth]{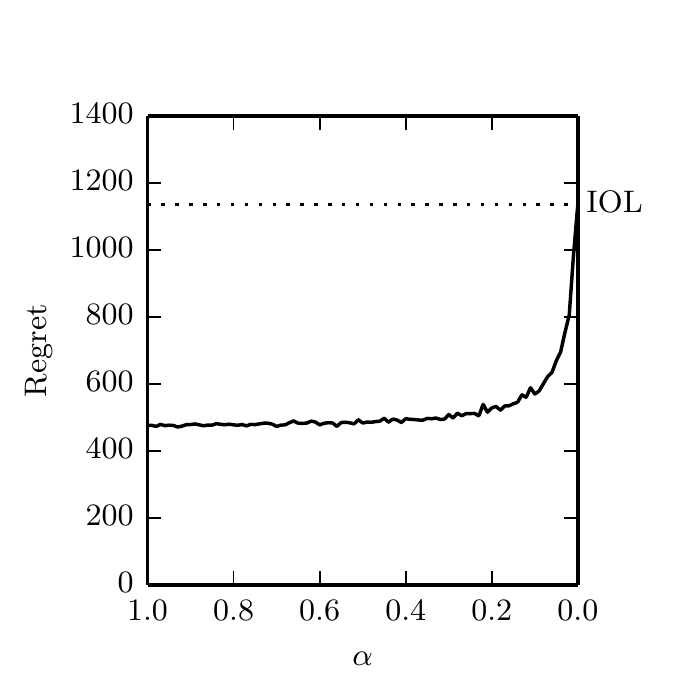}
    \label{fig.alpha}
   }   
   \subfigure[Varying $\beta$]{
     \includegraphics[trim = 2.5mm 3.2mm 4mm 10.5mm, clip=true, width=0.32\textwidth]{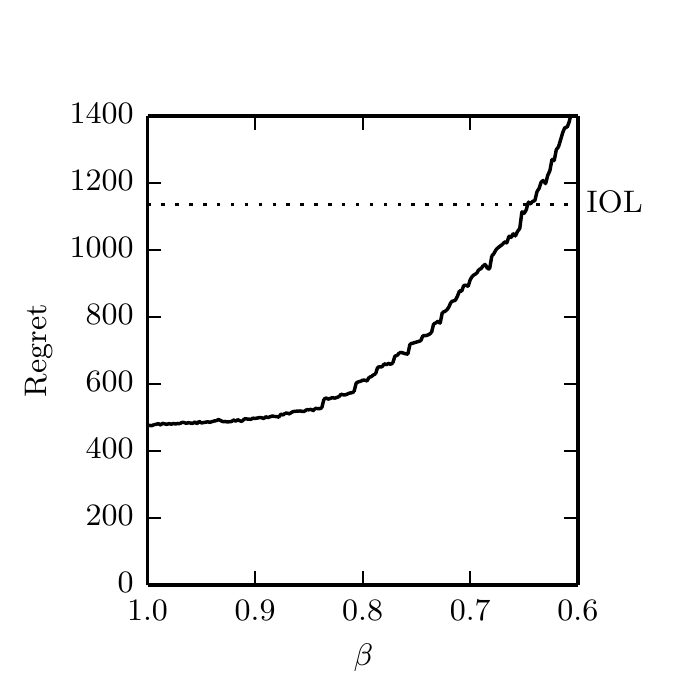}
     \label{fig.beta}
   } 
   \subfigure[Projection runtime varying $\beta$]{
     \includegraphics[trim = 2.5mm 3.2mm 4mm 10.5mm, clip=true, width=0.32\textwidth]{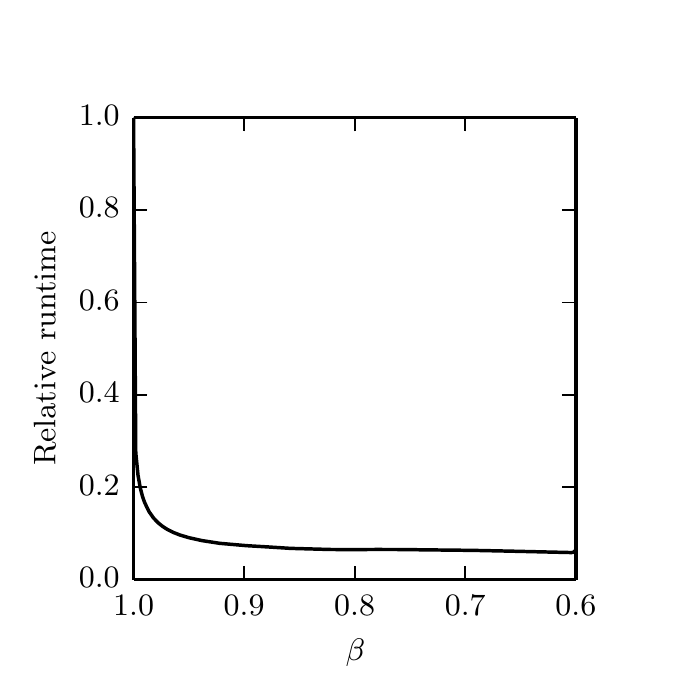}
     \label{fig.time}
   }    
\caption{Simulation results for learning hemimetrics. (a,b,c) compare the performance of \COCP against \IOCP for different orders of problem instances. (d,e,f)  show the speed/performance tradeoff using sporadic and approximate projections.}
\label{fig.simulations}
\vspace{-2mm}
\end{figure*}

In this section, we analyze worst-case regret bounds of the \COCP algorithm against a competing weight vector $\w^* \in S^*$. The proofs are provided in the extended version of this paper  \cite{hirnschall2017learning}.

\subsection{General Bounds}
We begin with a general result, without assumptions on the projection accuracy and rate.
\begin{theorem} \label{theorem:COCP}
The regret of the \COCP algorithm is bounded by $\regret_{\COCP}(T) \leq$
\begin{align}
&\quad \frac{1}{2 \eta} \norm{\solutionspace_{max}}^2 \sqrt{T \problems} +  2 \eta \norm{\maxgradient_{max}}^2 \sqrt{T\problems} \label{theorem:COCP.term1}  \tag{R1} \\
&\quad+  \sum^T_{t=1} \bad \norm{\solutionspace_{max}}  \norm{\maxgradient_{max}} \label{theorem:COCP.term2} \tag{R2} \\
&\quad+ \frac{1}{\eta} \sum^T_{t=1}  \inside \left( \maxdualitygap + \sqrt{2 \maxdualitygap}(t\problems)^{1/4} \norm{\solutionspace_{max}} \right) \label{theorem:COCP.term3} \tag{R3} \\
&\quad+ \frac{1}{2 \eta} \norm{\solutionspace_{max}}^2  - 2 \eta \norm{\maxgradient_{max}}^2 \problems. \label{theorem:COCP.term4} \tag{R4}
\end{align}
\end{theorem}
The regret in Theorem~\ref{theorem:COCP} has four components. \ref{theorem:COCP.term1} comes from the standard regret analysis in the OCP framework, \ref{theorem:COCP.term2} comes from sporadic projections, \ref{theorem:COCP.term3} comes from the allowed error in the projections, and \ref{theorem:COCP.term4} is a constant.

Note that when $\xi^t = 0$ for all $t$ (\emph{i.e.} no projections are performed) and $\eta$ is proportional to $\sfrac{1}{\sqrt{\counter^t_{i,j}}}$, we get the same regret bounds proportional to $\sqrt{T}$ as for the \IOCP algorithm. This also reveals the worst-case nature of the regret bounds of Theorem~\ref{theorem:COCP}, \emph{i.e.} the proven bounds for \COCP are agnostic to the specific structure $\solutionspace^*$ and the order of task instances. 

\subsection{Sporadic/Approximate Projection Bounds}
To provide specific bounds for the practically useful setting of sporadic and approximate projections, we introduce $\alpha$ and $\beta$ and the user chosen parameters $\calpha$ and $\cbeta$ to control the frequency and accuracy of the projections.
\begin{corollary} \label{corollary:COCP.rare}
Set $\eta = \frac{1}{2} \frac{\norm{\solutionspace_{max}}}{\norm{\maxgradient_{max}}}$. $\forall t \in [T]$, define
\begin{align*}
\xi^t \sim Bernoulli(\alpha) \textnormal{ with } \alpha = \frac{\calpha}{\sqrt{T}}, \\
\maxdualitygap = \cbeta (1 - \beta)^2 \frac{\sqrt{\problems}}{\sqrt{t}} \norm{\solutionspace_{max}}^2
\end{align*}
where constants $\calpha \in [0, \sqrt{T}]$, $\cbeta \geq 0$, and $\beta \in [0,1]$. The expected regret (w.r.t. $(\xi^t)_{t\in[T]}$) of the \COCP algorithm is bounded by $\expectedvalue{\regret_{\COCP}(T)} \leq$
\begin{align*}
&\quad 2 \sqrt{T \problems} \norm{\solutionspace_{max}}  \norm{\maxgradient_{max}} \cdot \bigg(1 + \frac{\calpha}{2\sqrt{\problems}}  \Big(1 - \frac{\calpha}{\sqrt{T}} \Big)& \\
& \qquad \qquad \qquad \qquad \qquad \quad + \calpha (\cbeta + \sqrt{2 \cbeta}) (1 - \beta) \bigg).&
\end{align*}
\end{corollary}

As shown in Corollary~\ref{corollary:COCP.rare}, projections are required to be more accurate for higher values of $t$. Intuitively, this is required so that already learned weights are not unlearned through inaccurate projections. Using the definitions under Corollary~\ref{corollary:COCP.rare}, we can prove worst-case regret bounds proportional to $\sqrt{T}$ for this setting.

\subsection{Performance Analysis for Hememtric Structure}
We now test the performance of the \COCP algorithm on synthetic data with an underlying hemimetric structure.

{\bfseries Hemimetric projection.} To be able to perform weighted projections onto the hemimetric polytope, we use the metric nearness algorithm \cite{sra2004triangle} as a starting point. For our purposes, three modifications of the algorithm are required: First, we lift the requirement of symmetry to generalize from metrics to hemimetrics. Second, the metric nearness algorithm does not guarantee a solution in the metric set in finite time. However, to calculate the duality gap, the solution is required to be feasible. Thus, we apply the Floyd-Warshall algorithm \cite{floyd1962algorithm} after every iteration to receive a solution in the hemimetric set. Third, we add weights to the triangle inequalities to allow for weighted projections and further add upper and lower bound constraints.

\begin{figure*}[!t]
\centering
     \includegraphics[width=1\textwidth]{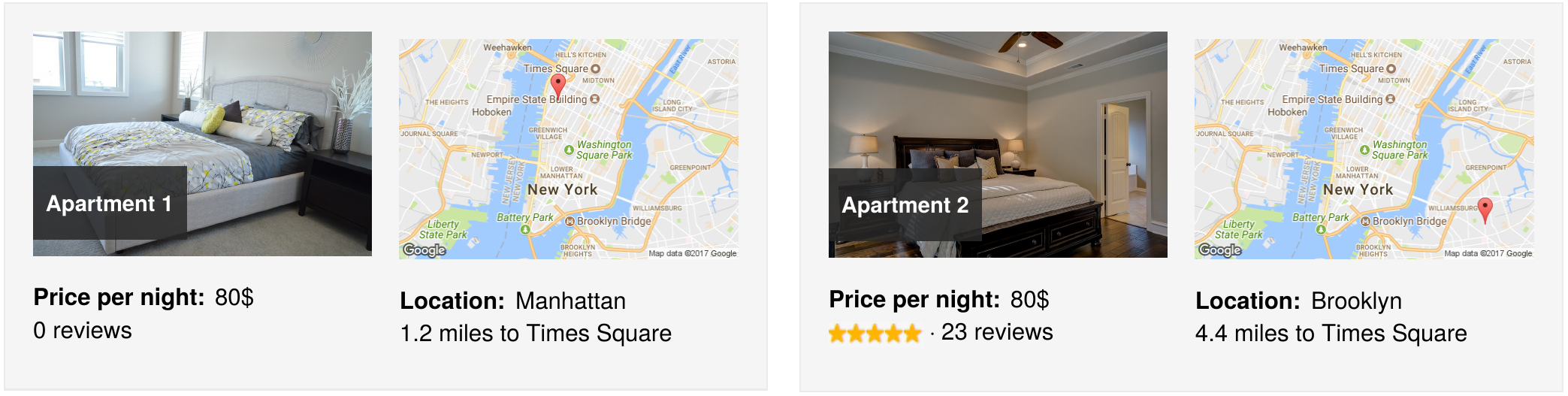}
\caption[Snapshot]{Snapshot of the user survey study on MTurk.\protect\footnotemark}
\label{fig.setup}
\vspace{-2mm}
\end{figure*}

{\bfseries Data structure.}  To empirically test the performance of the \COCP algorithm on the hemimetric set, we synthetically generate data with $d=1$ and model the underlying structure $\solutionspace^*$ as a set of $r$-bounded hemimetrics with $n=10$, resulting in $\problems = 90$ problems. We use a simple underlying \emph{ground-truth} hemimetric $\w^*$, where the $n$ items belong to two equal-sized clusters, with $p^*_{i,j} = 1$ if $i$ and $j$ are from the same cluster and $p^*_{i,j} = 9$ otherwise. The results of our experiment in Figure~\ref{fig.simulations} illustrate the potential runtime improvement using sporadic/approximate projections.

{\bfseries Random order of problems.} Problem instances $\z^t$ are chosen uniformly at random at every time step. The \COCP algorithm achieves a significantly lower regret than the \IOCP algorithm, benefiting from the weighted projections onto $\solutionspace^*$. At $T=500$, the regret of \COCP is less than half of that of the \IOCP, \emph{cf.} Figure~\ref{fig.random}.

{\bfseries Batches of problems.} In the batch setting, a problem instance is chosen uniformly at random, then it is repeated five times before choosing a new problem instance. Compared to the above-mentioned random order, the \IOCP algorithm suffers a lower regret because of a higher probability that problems are repeatedly shown. Furthermore, the benefit of the projections onto $\solutionspace^*$ for the  \COCP algorithm is reduced, \emph{cf.} Figure~\ref{fig.batch}, showing that the benefit of the projections depends on the specific order of the problem instances for a given structure.

\footnotetext{Real pictures from \names{Airbnb} replaced with illustrative examples.}

{\bfseries Single-problem setting.} A single problem $\z$ is repeated in every round. As illustrated, in this case the \IOCP algorithm and the \COCP algorithm have the same regret, \emph{cf.} Figure~\ref{fig.single}. In order to get a better understanding of using weights $Q^t$ for the weighted projection, we also show a variant uw-\COCP using $Q^t$ as identity matrix. Unweighted projection or using the wrong weights can hinder the convergence of the learners, as shown in Figure~\ref{fig.single} for this extreme case of a single-problem setting.

{\bfseries Varying the rate of projection (\boldmath$\alpha$).} The regret of the \COCP algorithm monotonically increases as $\alpha$ decreases, and is equivalent to the regret of the \IOCP algorithm at $\alpha=0$, \emph{cf.} Figure~\ref{fig.alpha}. In the range of $\alpha$ values between $1$ and $0.1$, the regret of the \COCP algorithm is relatively constant and increases strongly only as $\alpha$ approaches $0$. With $\alpha$ as low as $0.1$, the regret of the \COCP algorithm in this setting is still almost half of that of the \IOCP algorithm. 

{\bfseries Varying the accuracy of projection (\boldmath$\beta$).} The regret of the \COCP algorithm monotonically increases as $\beta$ decreases, and exceeds that of the \IOCP algorithm for values smaller than $0.65$ because of high errors in the projections, \emph{cf.} Figure~\ref{fig.beta}. In the range of $\beta$ values between $1$ and $0.85$, the regret of the \COCP algorithm is relatively constant and less than half of that of the \IOCP algorithm.

{\bfseries Runtime vs. approximate projections.} As expected, the runtime of the projection monotonically decreases as $\beta$ decreases, \emph{cf.} Figure~\ref{fig.time}. For values of $\beta$ smaller than $0.95$, the runtime of the projection is less than $10\%$ of that of the exact projection. Thus, with $\beta$ values in the range of $0.85$ to $0.95$, the \COCP algorithm achieves the best of both worlds: the regret is significantly smaller than that of \IOCP, with an order of magnitude speed up in the runtime compared to exact projections.

%%%%%%%%%%%%%%%%%%%%%%%%%%%%%%%%%%%%%%%%%%%%%%%%%%%%%%%%%%%%
%%%%%%%%%%%%%%%%%%%%%%%%%%%%%%%%%%%%%%%%%%%%%%%%%%%%%%%%%%%% CASE STUDY
\section{\names{Airbnb} Case Study}\label{sec.userstudy}

\begin{figure*}[!t]
\centering
   \subfigure[Dataset of 20 apartments]{
     \includegraphics[trim = 0mm 0mm 0mm 25mm, clip=true, width=0.3\textwidth]{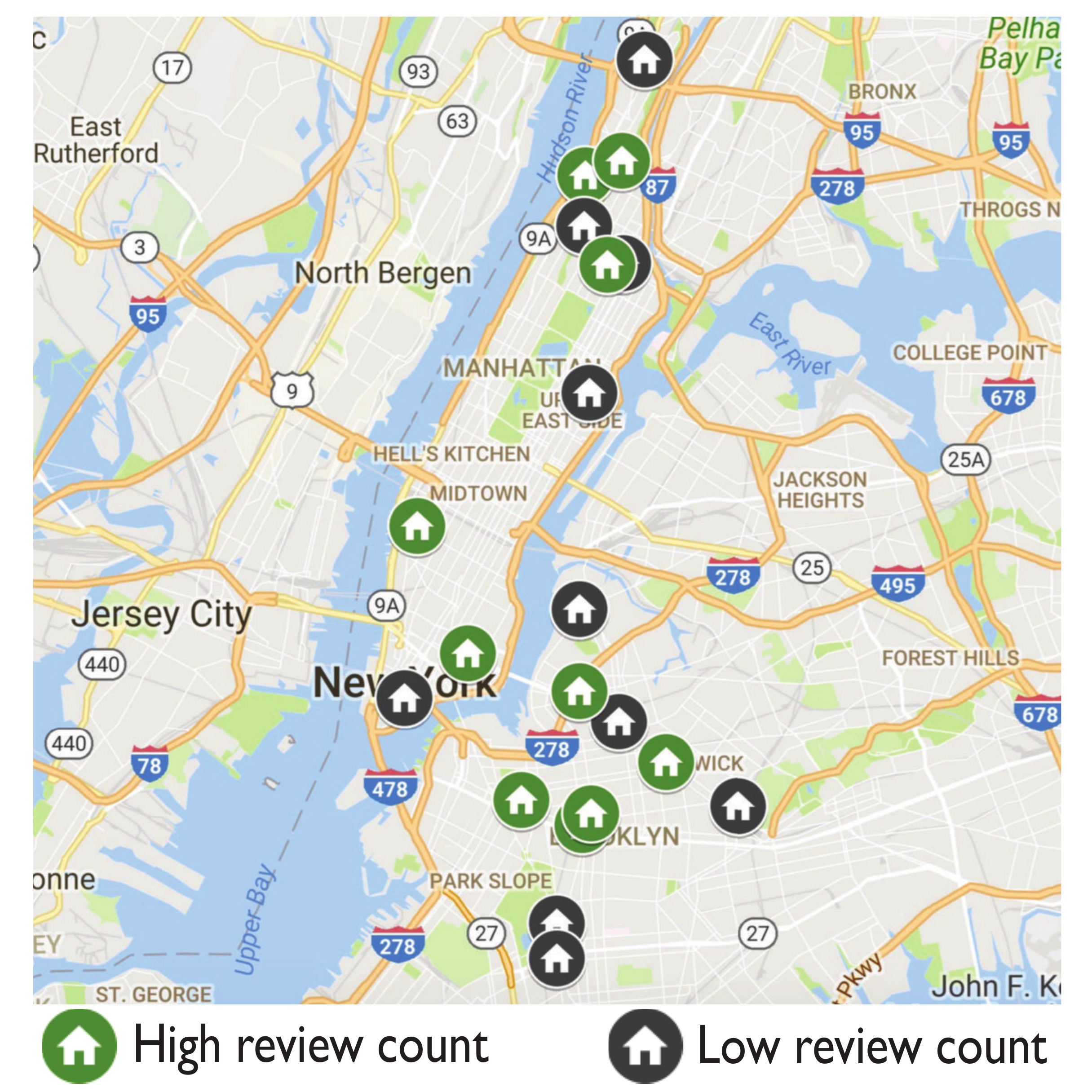}
    \label{fig.map}
   }   
   \subfigure[Cummulative utility gain]{
     \includegraphics[trim = 2.6mm 3.2mm 1.5mm 2mm, clip=true, width=0.32\textwidth]{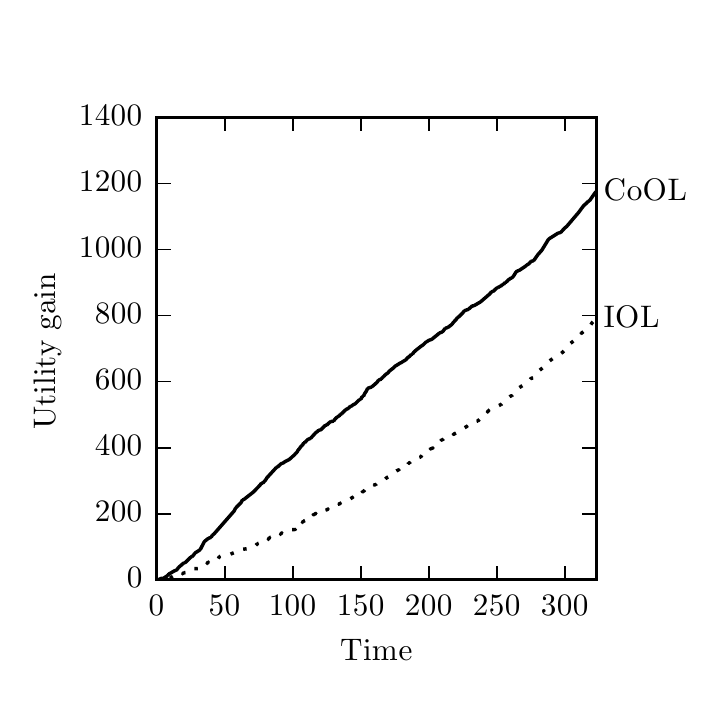}
     \label{fig.cummulative_reward}
   } 
   \subfigure[Average utility gain]{
     \includegraphics[trim = 2.6mm 3.2mm 1.5mm 2mm, clip=true, width=0.32\textwidth]{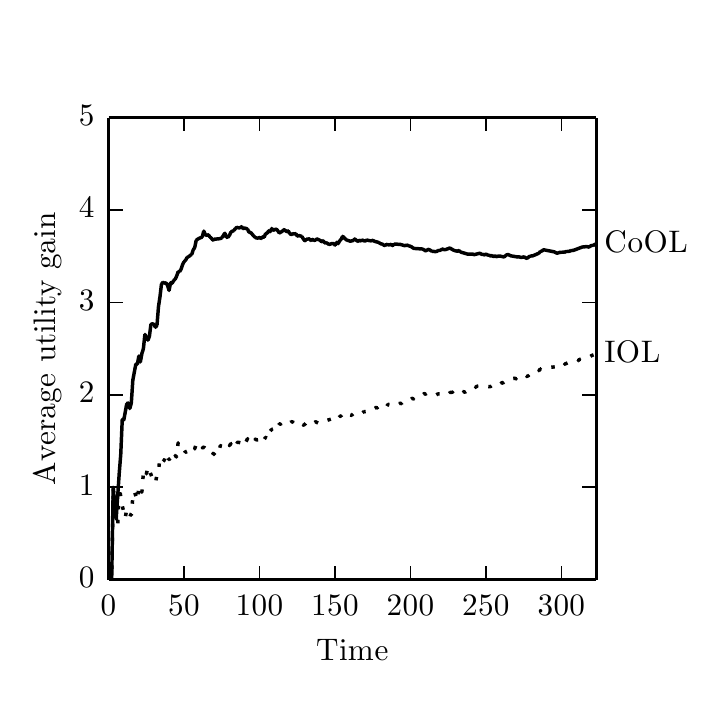}
     \label{fig.user_highlow_avg}
   }  
\caption{Results of experiments with \names{Airbnb} dataset.}
\label{fig.realworld}
\vspace{-2mm}
\end{figure*}

To test the viability and benefit of the \COCP algorithm in a realistic setting, we conducted a user study with data from the marketplace \names{Airbnb}.

\subsection{Experimental Setup}
We use the following setup in our user study:

{\bfseries \names{Airbnb} dataset.}
Using data of \names{Airbnb} apartments from \url{insideairbnb.com}, we created a dataset of 20 apartments as follows: we chose apartments from $4$ types in New York City by location (Manhattan or Brooklyn) and number of reviews (high, $\geq 20$ or low, $\leq 2$). From each type we chose 5 apartments, resulting in a total sample of $n=20$ apartments.

{\bfseries Survey study on MTurk platform.} In order to obtain real-world distributions of the users' private costs, we collected data from Amazon's Mechanical Turk marketplace. After several introductory questions about their preferences and familiarity with travel accommodations, participants were shown two randomly chosen apartments from the \names{Airbnb} dataset. To choose between the apartment, participants were given the price, location, picture, number of reviews and rating of each apartment, as shown in Figure~\ref{fig.setup}. Participants were first asked to select their preferred choice between the two apartments. Next, they were asked to specify their private cost for choosing the other, less preferred apartment instead. The collected data from the responses consists of tuples $((i,j),c)$, where $i$ is the preferred choice, $j$ is the suggested alternative, and $c$ is the private cost of the user. 

{\bfseries Sample.} In total, we received $943$ responses, as summarized in Table~\ref{table:surveyresults}. The sample for the performance analysis of the \COCP algorithm consists of $323$ responses, in which $i$ was a frequently reviewed apartment, $j$ an infrequently reviewed apartment, and participants were willing to explore the infrequently reviewed apartment for a discount (\emph{i.e.} they did not select \textnormal{NA}). 

\begin{table}[b]
\vspace{-2mm}
\centering
\renewcommand{\arraystretch}{1.3}
\begin{tabulary}{\linewidth}{C C C C C}
 \hline
  $i$ & $j$ & Responses & Accepted & Avg. Discount\\
 \hline
 High & Low         &416    &    77.6\%    & 29.5\$\\
 Low & Low         &228    &    83.3\%     & 28.1\$\\
 High & High     &219    &    82.2    \%    & 25.4\$\\
 Low & High         &80    &    81.3\%    & 25.9\$\\
 \hline
 \end{tabulary}  
\caption[]{Responses for different apartment types.} 
  \label{table:surveyresults}
\end{table}

{\bfseries Utility gain.}
The utility gain $u$ for getting a review for infrequently reviewed apartments is set to $u=40$ in our experiments, based on referral discounts given by \names{Airbnb} in the past.

{\bfseries Loss function.} As introduced in the methodology section, we require a convex version of the true loss function for our online learning framework, ideally acting as a surrogate of the true loss. Additionally, the gradient of the loss function needs to be calculated from the binary feedback of acceptance/rejection of the offers. However, in the analyzed model with binary feedback, a loss function that satisfies both requirements cannot be constructed. Instead, we consider a simplified piece-wise linear convex loss function given by $l^t(p^t) = \mathds{1}_{\{p^t \geq c^t\}} \cdot (p^t - c^t) + \mathds{1}_{\{p^t < c^t\}} \cdot \frac{u}{\Delta} \cdot (c^t - p^t)$, where $\frac{u}{\Delta}$ denotes the magnitude of the gradient when a user rejects the offer. For the experiment, we use a delta value of $20$. Due to this transformation, we use the utility gain rather than the loss as a useful measure of the performance of the \COCP algorithm. 

{\bfseries Structure.} Due to the small number of apartments, we consider a non-contextual setting with $d=1$ and use an r-bounded hemimetric structure to model the relationship of the tasks, where $r$ is set to $40$ to avoid recommending incentives $\prediction^t > \utility$. Using a setting with $d>1$ would allow for real-world applications with additional context.

\begin{table}[b]
\vspace{-2mm}
\centering
\renewcommand{\arraystretch}{1.3}
\begin{tabular}{l l r}
 \hline
  Category    \qquad \qquad    &  Example keywords   \qquad  \quad &      Mentions\\
 \hline
 Location     &  neighborhood, distance     &477\\
 Reviews     & rating, star                 &309 \\
 Price         &     expensive, cheap        & 182 \\
 Picture     &  image, photo                & 169 \\
 \hline
 \end{tabular}  
\caption[]{Mentioned categories for deciding on a discount.} 
  \label{table:survekeywords}
\end{table}

\subsection{Main Results}

We now present and discuss the results of the user study.

{\bfseries Descriptive statistics.} Out of all responses, 758 (80.4\%) respondents were willing to accept an offer for their less preferred apartment, given a certain discount per night.
Out of these respondents, the average required discount for accepting the alternative apartment was 27.9 USD per night. The average required discount for switching from a frequently reviewed apartment to an infrequently reviewed apartment was  $6\%$ higher.

Out of the respondents who could choose between a frequently and an infrequently reviewed apartment, 83.9\% respondents chose the frequently reviewed apartment, while only 16.1\% respondents chose the infrequently reviewed apartment.

In the responses to an open question about the factors respondents considered to decide on the discount, we captured the frequency at which different factors were mentioned by defining several keywords for each factor. The number of times each factor was mentioned is shown in Table~\ref{table:survekeywords}.

{\bfseries Algorithm performance.} We use the cumulative utility gain to measure the performance of the \IOCP and the \COCP algorithm. The utility gains of both algorithms after $323$ responses are shown in Figure~\ref{fig.user_highlow_avg}. The utility gain in Figure~\ref{fig.cummulative_reward} is almost 50\% higher for the \COCP algorithm than for the \IOCP algorithm. Figure~\ref{fig.user_highlow_avg} reveals that this gain is mainly achieved due to a significant speed up in learning over the first 50 problems.

{\bfseries Discussion.}
The results of the user study confirm several findings of \cite{fradkin2014search}, who studied the booking behavior on \names{Airbnb}. Similar to this study, we find that apartments with a high number of reviews are significantly more likely to be selected. We also find that the average required discount per night is higher when the alternative choice is an infrequently reviewed apartment. This also points toward a difference in willingness to pay between frequently and infrequently reviewed apartments. Similar results have been found in earlier studies on other marketplaces \cite{resnick2006value,ye2009impact,luca2011reviews}.

The user study also confirms that incentives influence buying behavior and can help increase exploration on online marketplaces \cite{avery1999market,robinson2012methods}; when respondents chose a frequently reviewed apartment and were asked to instead choose an infrequently reviewed apartment, 77.6\% of respondents were willing to accept a sufficiently large offer. More than 10\% of those respondents were willing to accept a discount of 10 USD per night or less.

The performance of the \IOCP and \COCP algorithm in Figures~\ref{fig.cummulative_reward} and \ref{fig.user_highlow_avg} suggests that incentives can be learned via online learning, and that structural information can be used to significantly speed up the learning. Further, the speed up in learning directly increases the marketplace's utility gain from suggesting alternative items. To reduce the problem size on a real-world application such as Airbnb, items could be grouped by features such as location or number of reviews. Further, problem-specific features, such as the distance between apartments could be added to increase the accuracy of the prediction.

%%%%%%%%%%%%%%%%%%%%%%%%%%%%%%%%%%%%%%%%%%%%%%%%%%%%%%%%%%%
%%%%%%%%%%%%%%%%%%%%%%%%%%%%%%%%%%%%%%%%%%%%%%%%%%%%%%%%%%% RELATED WORK
\section{Related Work}\label{sec.related}

{\bfseries Multi-armed bandit / Bayesian games.} A related path of research are multi-armed bandit and Bayesian games, where a principal attempts to coordinate agents to maximize its utility. Research in this area mainly focuses on changing the behavior of agents in the way information is disclosed, rather than through provision of payments. \cite{kremer2014implementing} provide optimal information disclosure policies for deterministic utilities and only two possible actions. \cite{mansour2015bayesian} generalize the results for stochastic utilities and a constant number of actions. Further, \cite{mansour2016bayesian} consider the interaction of multiple agents, and \cite{chakraborty2017coordinated} analyze a multi-armed bandits in the presence of communication costs. Our problem is different to previous research in that utilities are not required to be stochastic, and additional structural information is available to the principal.

{\bfseries Recommender systems.} A different approach to encouraging exploration in online marketplaces are recommender systems, which are known to influence buyers' purchasing decisions and can be used to encourage exploration \cite{resnick1997recommender,senecal2004influence}. For example, $\epsilon$-greedy recommender systems recommend a product closest to the buyer's preferences with probability ($1 - \epsilon$) and a random product with probability $\epsilon$ \cite{ten2003exploration}. Such recommender systems can be extended using ideas studied in this paper.

{\bfseries Online/distributed multi-task learning.}
Multi-task learning has been increasingly studied in online and distributed settings recently. 
Inspired by wearable computing, a recent work by \cite{jin2015collaborating} studied online multi-task learning in a distributed setting. They considered a setup, where tasks arrive asynchronously, and the relatedness among the tasks is maintained via a correlation matrix.  However, there is no theoretical analysis on the regret bounds for the proposed algorithms. \cite{wang2016distributed} recently studied the multi-task learning for distributed LASSO with shared support. Their work is different from ours --- we consider general convex constraints to model task relationships and consider the adversarial online regret minimization framework.

%%%%%%%%%%%%%%%%%%%%%%%%%%%%%%%%%%%%%%%%%%%%%%%%%%%%%%%%%%%
%%%%%%%%%%%%%%%%%%%%%%%%%%%%%%%%%%%%%%%%%%%%%%%%%%%%%%%%%%% CONCLUSION
\section{Conclusions and Future Work}\label{sec.conclusions}
We highlighted the need in the sharing economy to actively shape demand by incentivizing users to differ from their preferred choices and explore different options instead. To learn the incentives users require to choose different items, we developed a novel algorithm, \COCP, which uses structural information in user preferences to speed up learning. The key idea of our algorithm is to exploit structural information in a computationally efficient way by performing \emph{sporadic} and \emph{approximate} projections. We formally derived no-regret bounds for the \COCP algorithm and provided evidence for the increase in performance over the \IOCP baseline through several experiments. In a user study with apartments from the rental marketplace \names{Airbnb}, we demonstrated the practical applicability of our approach in a real-world setting. To conclude, we discuss several additional considerations for offering incentives in a sharing economy platform.

{\bfseries Safety/individual consumer loss.} Generally, exploration in the sharing economy may be risky, and individuals can face severe losses while exploring. For example, new hosts might not be trustworthy, and new drivers in ridesharing systems might not be reliable. In our approach, the items to be explored are controlled by the platform, and appropriate preconditions would need to be implemented to minimize risks.

{\bfseries Reliability/Consistency.} In order for platforms to implement an algorithmic provision of monetary incentives, it is important that incentives are reliable and consistent over time. Ideally, similar users should receive similar incentives, and offers should be consistent with the user's preferences. Using the \COCP algorithm, consistency can be controlled through appropriate convex constraints.

{\bfseries Strategy-proofness.} Providing monetary incentives based on user preferences creates possibilities for opportunistic behavior. For example, users could attempt to repeatedly decline offers to receive higher offers in the future or browse certain items hoping to receive offers for similar items. To control for such behavior, markets need to be large enough so that behavior of individuals does not affect overall learning. Further, platforms can control the number and frequency with which individual users receive offers to minimize opportunistic possibilities.

\subsection{Acknowledgments} This work was supported in part by the Swiss National Science Foundation, and Nano-Tera.ch program as part of the Opensense II project, ERC StG 307036, and a Microsoft Research Faculty Fellowship. Adish Singla acknowledges support by a Facebook Graduate Fellowship.

%%%%%%%%%%%%%%%%%%%%%%%%%%%%%%%%%%%%%%%%%%%%%%%%%%%%%%%%%%
%%%%%%%%%%%%%%%%%%%%%%%%%%%%%%%%%%%%%%%%%%%%%%%%%%%%%%%%%% BIBLIOGRAPHY
\bibliography{hirnschall-singla}
\bibliographystyle{aaai}

%!TEX root = main.tex

\onecolumn
\appendix
{\allowdisplaybreaks

\section{Outine of the Supplement}
We start the supplement by introducing properties of the Bregman divergence and additional notation required for the proofs of the regret bounds. We further introduce two basic propositions and several lemmas. We then provide formal justification for using weighted projection in the \COCP algorithm, \emph{cf.} Equation~\eqref{eq.weightedproj}. Lastly, we provide the proof of the regret bound of the \COCP algorithm in Theorem~\ref{theorem:COCP} and Corollary~\ref{corollary:COCP.rare}.

%!TEX root = ../main.tex

\section{Preleminaries} \label{appendix_bregman-divergences}
\subsection{Bregman Divergence}
For any strictly convex function $\regularizer^t: \mathbb{R}^d \to \mathbb{R}$, the Bregman divergence $\divergence_{\regularizer^t}$ between $\pmb{a}$, $\pmb{b} \in \mathbb{R}^d$ is defined as the difference between the value of $\regularizer^t$ at $\pmb{a}$, and the first-order Taylor expansion of $\regularizer^t$ around $\pmb{b}$ evaluated at $\pmb{a}$, i.e.
$$\divergence_{\regularizer^t}(\pmb{a}, \pmb{b}) = \regularizer^t(\pmb{a}) - \regularizer^t(\pmb{b}) - \nabla \regularizer^t(\pmb{b}) \cdot (\pmb{a} - \pmb{b}), $$

We use the following properties of the Bregman divergence, \emph{cf.} \cite{rakhlin2009lecture}:
\begin{itemize}
\item The Bregman divergences is non-negative.
\item The Bregman projection $$\widehat{\pmb{b}} = \argmin_{\pmb{a} \in \solutionspace} \divergence_{\regularizer^t}(\pmb{a}, \pmb{b})$$
onto a convex set $\solutionspace$ exists and is unique.
\item For $\widehat{\pmb{b}}$ defined as in the Bregman projection above and $\pmb{u} \in \solutionspace$, by the generalized Pythagorean theorem,  \emph{cf.} \cite{cesa2006prediction}, the Bregman divergence satisfies
 $$\divergence_{\regularizer^t}(\pmb{u}, \pmb{b}) \geq \divergence_{\regularizer^t}(\pmb{u}, \widehat{\pmb{b}}) + \divergence_{\regularizer^t}(\widehat{\pmb{b}}, \pmb{b}).$$
\item The three-point equality 
$$\divergence_{\regularizer^t}(\pmb{a}, \pmb{b}) + \divergence_{\regularizer^t}(\pmb{b}, \pmb{c}) = \divergence_{\regularizer^t}(\pmb{a}, \pmb{c}) + (\pmb{a} - \pmb{b}) (\nabla \regularizer(\pmb{c}) - \nabla \regularizer(\pmb{b}))$$

follows directly from the definition of the Bregman divergence.
\end{itemize}

\subsection{Notation}
Throughout the supplement we use  $\eta^t_\z = \frac{\eta}{\sqrt{\counter^t_\z}}$ and $\Q^t$ as per Equation~\eqref{eq.qmatrix}. Similar to the definition of $\w^t$, we also define $\x^t$, and $\gradient^t$ as the concatenation of the task specific feature and gradient vectors, \emph{i.e.}
$$\x^t=\big[(\x^t_1)' \ {\cdots} (\x^t_\z)' \ {\cdots} (\x^t_\problems)'\big]'
\qquad
\gradient^t=\big[(\gradient^t_1)' \ {\cdots} (\gradient^t_\z)' \ {\cdots} (\gradient^t_\problems)'\big]'.$$

where for all $t$, $\x^t$ and $\gradient^t$ are $0$ in all positions that do not correspond to task $\z^t$. We also use $\wtilde^{t+1}$ to refer to the concatenation of the updated task specific weights, before any coordination, such that
$$\wtilde^{t+1}=\big[(\w_1)' \ {\cdots} (\w_\z)' \ {\cdots} (\w_\problems)'\big]'.$$

where $\w_z = \w^t_\z$ for $\z \neq z^t$ and $\w_\z = \wtilde^t_z$ otherwise.

%!TEX root = ../main.tex

\section{Propositions}\label{appendix_propositions}
In the following we introduce two basic propositions that we need for the proof of Theorem~\ref{theorem:COCP}.

\begin{proposition} \label{proposition:tz}
If $\counter_\z \in \mathbb{R}^+$ for all $\z \in \{1 \dots \problems\}$, and $\sum\limits^\problems_{\z=1} \counter_\z = T ,$
then
$$\sum^\problems_{\z=1} \sqrt{\counter_\z} \leq \sqrt{T \problems} .$$
\end{proposition}

\begin{proof}

Extending and applying the Cauchy-Schwarz inequality, we get
\begin{align*}
\sum^\problems_{\z=1} \sqrt{\counter_\z} &\leq \sqrt{\sum^\problems_{\z=1} 1}\sqrt{\sum^\problems_{\z=1} \counter_\z} \\
&= \sqrt{\problems} \sqrt{T} \\
&= \sqrt{T \problems}
\end{align*}

\end{proof} 

\begin{proposition} \label{proposition:sumt}
The sum from $\sum^T_{t=1}\frac{1}{\sqrt{t}}$ is bounded by $2 \sqrt{T} - 1$.
\end{proposition}
\begin{proof}
\begin{align*}
\sum^T_{t=1}\frac{1}{\sqrt{t}} &\leq 1 + \int^T_{t=1} \frac{1}{\sqrt{t}} dt \\
&= 1 + \left[2 \sqrt{t} \right]^T_1 \\
&= 2 \sqrt{T} - 1
\end{align*}
\end{proof}

%!TEX root = ../main.tex

\section{Lemmas}\label{appendix_lemmas}
In this section we introduce the lemmas required for the proof of the regret bounds of the \COCP algorithm. Applying Lemma~\ref{lemma:linearloss} allows us to replace the loss function with its linearization, similar to \cite{zinkevich2003online}. Lemmas~\ref{lemma:update1} and~\ref{lemma:update2} allow us to get an equivalent update procedure, using the Bregman divergence, and Lemma~\ref{lemma:constraint-minimizer} gives a handle on the linearized regret bound, \emph{cf.} \cite{rakhlin2009lecture}.  Lemma~\ref{lemma:dualitygap} uses the duality gap to upper bound the Bregman divergence between the exact and approximate projection. Lemmas~\ref{lemma:maxdivergence} and~\ref{lemma:rtplusminusrt} provide different upper bounds on the Bregman divergence.

\begin{lemma} \label{lemma:linearloss}
For all $t$ and $\w^t_\z$ there exists a $\gradient^t_\z \in \mathbb{R}^d$ such that  $\loss^t(\w^t_\z)$ can be replaced with $\gradient^t \cdot \w^t_\z$ without loss of generality.
\end{lemma}

\begin{proof}
The loss function affects the regret in two ways: First, the loss function's gradient is used in the update step, and second, the loss function is used to calculate the regret of the algorithm. Let $\gradient^t_\z = \nabla \loss^t(\w^t_\z)$ and consider the linearized loss $\gradient^t_\z \cdot \w^t_\z$. Using the linearized loss, the behavior of the algorithm remains unchanged, since $\nabla \loss^t(\w^t_\z) = \gradient^t_\z$. Further, the regret either increases or remains unchanged, since the loss function is convex, such that for all $\wstar_\z \in \solutionspace_\z$
$$\loss^t(\wstar_\z) \geq \gradient^t_\z \cdot (\wstar_\z - \w^t_\z) + \loss^t(\w^t_\z). $$

Rearranging, we get
\begin{align*}
\loss^t(\w^t_\z) - \loss^t(\wstar_\z) & \leq \gradient^t_\z \cdot \w^t_\z - \gradient^t_\z \cdot \wstar_\z,
\end{align*}

such that using a linearized loss, the regret either remains constant or increases.
\end{proof}

\begin{lemma} \label{lemma:update1}
For $\regularizer^t(\w) = \frac{1}{2} \w' \Q^t \w$, the update rule
$$\wtilde^{t+1} = \w^t - \eta^t_\z \gradient^t$$
is equivalent to the update rule
$$\wtilde^{t+1} = \argmin_{\w \in \mathbb{R}^{d \problems}} \eta \gradient^t \cdot \w + \divergence_{\regularizer^t}(\w, \w^t).$$
\end{lemma}
\begin{proof}
For the second update rule, inserting $\regularizer^t(\w) = \frac{1}{2} \w' \Q^t \w$ into the definition of the Bregman divergence and setting the derivative with respect to $\w$ evaluated at $\wtilde^{t+1}$ to zero, we have
$$\eta \gradient^t + \wtilde^{t+1} \Q^t - \w^t \Q^t = 0$$

Rewriting, using that $\gradient^t$ is non-zero only in entries that correspond to $\z^t$, and applying the definitions of $\Q^t$ and $\eta$, we get
\begin{align*}
\wtilde^{t+1} &= \w^t - \eta \gradient^t (\Q^t)^{-1} \\
&= \w^t - \frac{\eta}{\sqrt{\counter^t_\z}} \gradient^t \\
&= \w^t - \eta^t_\z \gradient^t.
\end{align*}
\end{proof}

\begin{lemma} \label{lemma:update2}
For $\regularizer^t(\w) = \frac{1}{2} \w' \Q^t \w$, the update rule
$$\w^{t+1} = \argmin_{\w \in \solutionspace^*} \divergence_{\regularizer^t}(\w, \wtilde^{t+1}),$$
where $ \wtilde^{t+1} = \w^t - \eta^t_\z \gradient^t$,
is equivalent to the update rule
$$\w^{t+1} = \argmin_{\w \in \solutionspace^*} \eta \gradient^t \cdot \w + \divergence_{\regularizer^t}(\w, \w^t)$$
\end{lemma}
\begin{proof}

Applying the definition of $\regularizer^t(\w)$, we can rewrite 
\begin{align*}
\w^{t+1} &= \argmin_{\w \in \solutionspace^*} \divergence_{\regularizer^t}(\w, \wtilde^{t+1}) \\
&= \argmin_{\w \in \solutionspace^*} \frac{1}{2} (\w - \w^t + \eta \gradient^t (\Q^t)^{-1})' \Q^t (\w - \w^t + \eta \gradient^t (\Q^t)^{-1}) \\
&= \argmin_{\w \in \solutionspace^*} \eta \gradient^t \cdot \w  + \frac{1}{2} (\w - \w^t)' \Q^t (\w - \w^t) \\
&= \argmin_{\w \in \solutionspace^*} \eta \gradient^t \cdot \w + \divergence_{\regularizer^t}(\w, \w^t)
\end{align*}
\end{proof}

\begin{lemma} \label{lemma:constraint-minimizer}
If $\w^{t+1}$ is the constraint minimizer of the objective $\eta \gradient^t \cdot \w + \divergence_{\regularizer^t}(\w, \w^t)$ as stated in Lemma \ref{lemma:update2}, then for any \pmb{a} in the solution space,
$$\eta \gradient^t \cdot (\w^{t+1} - \pmb{a}) \leq \divergence_{\regularizer^t}(\pmb{a}, \w^t) - \divergence_{\regularizer^t}(\pmb{a}, \w^{t+1}) - \divergence_{\regularizer^t}(\w^{t+1}, \w^t). $$
\end{lemma}
\begin{proof}
Since $\w^{t+1}$ is the constraint minimizer of the objective $\eta \gradient^t \cdot \w + \divergence_{\regularizer^t}(\w, \w^t)$, any vector pointing away from $\w^{t+1}$ into the solution space has a positive product with the gradient of the objective at $\w^{t+1}$, such that
$$ 0 \leq (\pmb{a} - \w^{t+1}) \cdot (\eta \gradient^t + \nabla \regularizer^t(\w^{t+1}) -\nabla \regularizer^t(\w^t)). $$
Rewriting and using the three-point equality, we get
\begin{align*}
\eta \gradient^t \cdot (\w^{t+1} - \pmb{a}) &\leq (\pmb{a} - \w^{t+1}) \cdot (\nabla \regularizer^t(\w^{t+1}) - \nabla \regularizer^t(\w^t)) \\
&= \divergence_{\regularizer^t}(\pmb{a}, \w^t) - \divergence_{\regularizer^t}(\pmb{a}, \w^{t+1}) - \divergence_{\regularizer^t}(\w^{t+1}, \w^t).
\end{align*}
\end{proof}

\begin{lemma} \label{lemma:dualitygap}
If $\what^{t+1}$ is the exact solution of 
$$\argmin_{\w \in \solutionspace^*} \divergence_{\regularizer^t}(\w, \wtilde^{t+1})$$
and $\w^{t+1} \in \solutionspace^*$ is an approximate solution with duality gap less than $\maxdualitygap$, then
$$\maxdualitygap \geq \divergence_{\regularizer^t}(\what^{t+1}, \w^{t+1}) .$$
\end{lemma}
\begin{proof}
The duality gap is defined as the difference between the primal and dual value of the solution. The dual value is upper bounded by the optimal solution and thus less than or equal to $\divergence_{\regularizer^t}(\what^{t+1}, \wtilde^{t+1})$. Thus, for the primal solution $\divergence_{\regularizer^t}(\w^{t+1}, \wtilde^{t+1})$ with duality gap less than $\maxdualitygap$, we have
$$\maxdualitygap \geq \divergence_{\regularizer^t}(\w^{t+1}, \wtilde^{t+1}) - \divergence_{\regularizer^t}(\what^{t+1}, \wtilde^{t+1}) $$
Note that $\what^{t+1}$ is the projection of $\wtilde^{t+1}$ onto $\solutionspace^*$ and $\w^{t+1} \in \solutionspace^*$. Thus, using the propertiesof the Bregman divergence we can apply the generalized Pythagorean theorem such that
$$\divergence_{\regularizer^t}(\w^{t+1}, \wtilde^{t+1}) \geq \divergence_{\regularizer^t}(\w^{t+1}, \what^{t+1}) + \divergence_{\regularizer^t}(\what^{t+1}, \wtilde^{t+1}) $$

Inserting into the above inequality we get the result.
\end{proof}

\begin{lemma} \label{lemma:maxdivergence}
For $\regularizer^t(\w) = \frac{1}{2} \w' \Q^t \w$ and $\pmb{a}$ and $\pmb{b} \in \solutionspace$,
$$\divergence_{\regularizer^t}(\pmb{a}, \pmb{b}) \leq \frac{1}{2} \norm{\solutionspace_{max}}^2 \sqrt{t \problems}  $$
\end{lemma}

\begin{proof}
Using the definition of $\Q^t$, noting that $\norm{\pmb{a}_\z - \pmb{b}_\z}^2 \leq \norm{\solutionspace_{max}}^2$, and applying Proposition~\ref{proposition:tz} we can write
\begin{align*}
\divergence_{\regularizer^t}(\pmb{a}, \pmb{b}) &= \frac{1}{2} (\pmb{a} - \pmb{b})' \Q^t (\pmb{a} - \pmb{b}) \\
&= \frac{1}{2} \sum^\problems_{\z=1} \norm{\pmb{a}_\z - \pmb{b}_\z}^2 \sqrt{\counter^t_\z} \\
&\leq \frac{1}{2} \norm{\solutionspace_{max}}^2 \sum^\problems_{\z=1} \sqrt{\counter^t_\z} \\
&\leq \frac{1}{2} \norm{\solutionspace_{max}}^2 \sqrt{t \problems} 
\end{align*}
\end{proof}

\begin{lemma} \label{lemma:rtplusminusrt}
For any two $\pmb{a}^t$, $\pmb{b}^t \in \solutionspace$,
$$\sum^T_{t=1} \divergence_{\regularizer^{t+1}}(\pmb{a}^t, \pmb{b}^t) - \divergence_{\regularizer^t}(\pmb{a}^t, \pmb{b}^t) \leq \frac{1}{2} \norm{\solutionspace_{max}}^2 \sqrt{T \problems} .$$
\end{lemma}
\begin{proof}
Applying our definition of $\regularizer^t$, we can rewrite 
$$ \sum^T_{t=1} \divergence_{\regularizer^{t+1}}(\pmb{a}^t, \pmb{b}^t) - \divergence_{\regularizer^t}(\pmb{a}^t, \pmb{b}^t) = \frac{1}{2} \sum^T_{t=1} (\pmb{a}^t - \pmb{b}^t)' (\Q^{t+1} - \Q^t) (\pmb{a}^t - \pmb{b}^t). $$

Note that 
$$ (\Q^{t+1} - \Q^t) =
\begin{bmatrix}
\sqrt{\counter^{t+1}_1} - \sqrt{\counter^t_1} & & 0  \\
& \ddots & \\
0 & & \sqrt{\counter^{t+1}_\problems} - \sqrt{\counter^t_\problems}
\end{bmatrix}. $$

Applying Proposition~\ref{proposition:tz} and using $\norm{\pmb{a}_\z - \pmb{b}_\z}^2 \leq \norm{\solutionspace_\z}^2 \leq \norm{\solutionspace_{max}}^2$  we get
\allowdisplaybreaks
\begin{align*}
\sum^T_{t=1} \divergence_{\regularizer^{t+1}}\norm{\pmb{a}^t, \pmb{b}^t} - \divergence_{\regularizer^t}(\pmb{a}^t, \pmb{b}^t) &= \frac{1}{2}  \sum^T_{t=1} \sum^\problems_{\z=1} \left(\sqrt{\counter^{t+1}_\z} - \sqrt{\counter^t_\z}\right) \norm{\pmb{a}^t_\z - \pmb{b}^t_\z}^2 \\
&\leq \frac{1}{2} \norm{\solutionspace_{max}}^2 \sum^\problems_{\z=1} \sum^T_{t=1} \left(\sqrt{\counter^{t+1}_\z} - \sqrt{\counter^t_\z} \right) \\
&= \frac{1}{2} \norm{\solutionspace_{max}}^2  \sum^\problems_{\z=1} \left(\sqrt{\counter^{T+1}_\z} - \sqrt{\counter^{1}_\z} \right) \\
&= \frac{1}{2} \norm{\solutionspace_{max}}^2  \left( \sum^\problems_{\z=1} \left(\sqrt{\counter^{T+1}_\z} \right) - 1  \right) \\
&= \frac{1}{2} \norm{\solutionspace_{max}}^2 \left( \sum^\problems_{\z=1} \left(\sqrt{\counter^T_\z} \right) + \sqrt{\counter^{T+1}_{\z^{T+1}}} - \sqrt{\counter^T_{\z^{T+1}}}  - 1  \right)\\
&= \frac{1}{2} \norm{\solutionspace_{max}}^2  \left( \sum^\problems_{\z=1} \left(\sqrt{\counter^T_\z} \right) + \sqrt{\counter^T_{\z^{T+1}} + 1} - \sqrt{\counter^T_{\z^{T+1}}}  - 1  \right) \\
&\leq \frac{1}{2} \norm{\solutionspace_{max}}^2  \sum^\problems_{\z=1} \sqrt{\counter^T_\z} \\
&\leq \frac{1}{2} \norm{\solutionspace_{max}}^2 \sqrt{T \problems} \\
\end{align*}
\end{proof}

%!TEX root = ../ec-incentives-exploration.tex

\section{Idea of Weighted Projections}\label{appendix_idea-weighted-projection}
The update in Algorithm~\ref{algo.COCP} line~\ref{alg1.update} can be equivalently written as 
$$\wtilde^{t+1} = \w^t - \eta^t_\z \gradient^t.$$

As shown in Lemma~\ref{lemma:update1}, we can rewrite this as
$$\wtilde^{t+1} = \argmin_{\w \in \mathbb{R}^{d \problems}} \eta \gradient^t \cdot \w + \divergence_{\regularizer^t}(\w, \w^t),$$
using the regularizer $\regularizer^t(\w) = \frac{1}{2} \w \cdot \Q^t \w$.

Intuitively, the \COCP algorithm restricts the solution to $\solutionspace^*$, such that the update can be rewritten as 
\begin{align*}
\w^{t+1} &= \argmin_{\w \in \solutionspace^*} \eta \gradient^t \cdot \w + \divergence_{\regularizer^t}(\w, \w^t) \\
&= \argmin_{\w \in \solutionspace^*} \eta \gradient^t \cdot \w  + \frac{1}{2} (\w - \w^t) \cdot \Q^t (\w - \w^t) \\
&= \argmin_{\w \in \solutionspace^*} \frac{1}{2} (\w - \w^t + \eta \gradient^t (\Q^t)^{-1}) \cdot \Q^t (\w - \w^t + \eta \gradient^t (\Q^t)^{-1}) \\
&= \argmin_{\w \in \solutionspace^*} \frac{1}{2} (\w - \wtilde^{t+1}) \cdot \Q^t (\w - \wtilde^{t+1}),
\end{align*}

which is equal to the weighted projection introduced in Equation~\eqref{eq.weightedproj}. Using weights defined by some other heuristics could in general lead to a higher regret. For instance, in Figure~\ref{fig.single} we show the increase in regret of the \COCP algorithm, when setting $\Q^t$ as the identity matrix.

%!TEX root = ../main.tex

\section{Proof of Theorem \ref{theorem:COCP}}\label{appendix_theorems_main}
In the following we provide the proof of Theorem~\ref{theorem:COCP}, using notation and results of the earlier sections of the supplement. Unlike earlier work (\emph{e.g.} \cite{zinkevich2003online,rakhlin2009lecture}), in our setting projections are allowed to be noisy and therefore, the solution may not be a constraint minimizer of the projection. Additionally, in our setting projection may occur only sporadically, and thus intermediary solutions may not be in $\solutionspace^*$. To keep track of whether projection occurred, we define indicator functions and handle the special case of projection at time $t$ without projection at time $t-1$ separately.

\begin{proof}{Proof of Theorem~\ref{theorem:COCP}}
\subsection*{Preparation}
We define $\what^t$ as the exact solution of the projection onto $\solutionspace^*$, such that
$$\what^t = \argmin_{\w \in \solutionspace^*} (\w - \wtilde^{t+1})' \Q^t (\w - \wtilde^{t+1}) .$$

Recall that $\xi^t$ is $1$ with probability $\alpha$ and $0$ with probability $(1 - \alpha)$. The algorithm projects onto $\solutionspace^*$ if $\xi^t = 1$ and onto $\solutionspace_\z$ if $\xi^t = 0$. We define the indicator functions
\begin{align*}
\inside = 
\left\{
	\begin{array}{ll}
		1 & \mbox{if } \xi^t = 1\\
		0 & \text{otherwise} .
	\end{array}
\right.
\end{align*}
and the inverse

\begin{align*}
\outside = 
\left\{
	\begin{array}{ll}
		1 & \mbox{if } \xi^t = 0\\
		0 & \text{otherwise} .
	\end{array}
\right.
\end{align*}

as well as
\begin{align*}
\bad = 
\left\{
	\begin{array}{ll}
		1 & \mbox{if } \xi^{t-1} = 0 \text{ and } \xi^t = 1 \\
		0 & \text{otherwise} .
	\end{array}
\right.
\end{align*}
and the inverse
\begin{align*}
\good = 
\left\{
	\begin{array}{ll}
		1 & \mbox{if } \xi^{t-1} = 1 \text{ or } \xi^t = 0 \\
		0 & \text{otherwise} .
	\end{array}
\right.
\end{align*}

Our goal is to upper bound the regret, which, using Lemma~\ref{lemma:linearloss}, we can write as $$ \regret_{\COCP}(T)  = \sum^T_{t=1} \gradient^t_\zt \cdot (\w^t_\zt - \wstar_\zt)  .$$ Using the definitions above, we rewrite
\begin{align*}
\sum^T_{t=1} \gradient^t_\zt \cdot (\w^t_\zt - \wstar_\zt) &= \sum^T_{t=1} \gradient^t \cdot (\w^t - \wstar) \\
&= \sum^T_{t=1} \gradient^t \cdot (\what^{t+1} - \wstar) + \sum^T_{t=1} \gradient^t \cdot (\w^t - \what^{t+1}) \\
&= \sum^T_{t=1} \inside \gradient^t \cdot (\what^{t+1} - \wstar) + \sum^T_{t=1} \outside \gradient^t \cdot (\what^{t+1} - \wstar) \\
&\phantom{=} + \sum^T_{t=1} \good \gradient^t \cdot (\w^t - \what^{t+1}) + \sum^T_{t=1} \bad \gradient^t \cdot (\w^t - \what^{t+1}) .
\end{align*}

and further upper bound each sum individually.

Throughout the proof, we use the Bregman divergence with the regularizer $\regularizer^t(\w) = \frac{1}{2} \w' \Q^t \w$, and apply Lemmas~\ref{lemma:update1} and~\ref{lemma:update2} to get an equivalent update procedure.

\subsection*{Step 1: First sum}
Applying Lemma~\ref{lemma:constraint-minimizer} with $\what^{t+1}$ as the constraint minimizer of the objective $\eta \gradient^t \cdot \w + \divergence_{\regularizer^t}(\w, \w^t)$ and $\wstar \in \solutionspace^*$, we have
$$ \eta \gradient^t \cdot (\what^{t+1} - \wstar) \leq \divergence_{\regularizer^t}(\wstar, \w^t) - \divergence_{\regularizer^t}(\wstar, \what^{t+1}) - \divergence_{\regularizer^t}(\what^{t+1}, \w^t) . $$

Summing over time,
\begin{align*}
\eta \sum^T_{t=1} \inside \gradient^t \cdot (\what^{t+1} - \wstar) &\leq \sum^T_{t=1} \inside \left( \divergence_{\regularizer^t}(\wstar, \w^t) - \divergence_{\regularizer^t}(\wstar, \what^{t+1}) - \divergence_{\regularizer^t}(\what^{t+1}, \w^t) \right) \\
&= \sum^T_{t=1} \left(\insidet{t+1} \divergence_{\regularizer^{t+1}}(\wstar, \w^{t+1}) - \inside \divergence_{\regularizer^t}(\wstar, \what^{t+1}) - \inside \divergence_{\regularizer^t}(\what^{t+1}, \w^t) \right) \\
& \phantom{\leq} + \insidet{1} \divergence_{\regularizer^1}(\wstar, \w^1) - \insidet{T+1}  \divergence_{\regularizer^{T+1}}(\wstar, \w^{T+1}) \\
&\leq \sum^T_{t=1} \left(\insidet{t+1} \divergence_{\regularizer^{t+1}}(\wstar, \w^{t+1}) - \inside \divergence_{\regularizer^t}(\wstar, \w^{t+1}) \right) \\
& \phantom{=} + \sum^T_{t=1} \inside \left( \divergence_{\regularizer^t}(\wstar, \w^{t+1}) - \divergence_{\regularizer^t}(\wstar, \what^{t+1}) \right) \\
& \phantom{=} + \insidet{1} \divergence_{\regularizer^1}(\wstar, \w^1) .
\end{align*}

In the following, we upper bound each term individually. For now we leave the first term  unchanged and provide an upper bound in step 3 by combining it with the results of step 2.

For the second term, we use that for our choice of $\regularizer$, the square root of the Bregman divergence is a norm and therefore satisfies the triangle inequality. Thus,
$$ \sqrt{\divergence_{\regularizer^t}(\wstar, \w^{t+1})} \leq \sqrt{\divergence_{\regularizer^t}(\wstar, \what^{t+1})} + \sqrt{\divergence_{\regularizer^t}(\what^{t+1}, \w^{t+1})} .$$

Squaring both sides, we have
$$\divergence_{\regularizer^t}(\wstar, \w^{t+1}) \leq \divergence_{\regularizer^t}(\wstar, \what^{t+1}) + \divergence_{\regularizer^t}(\what^{t+1}, \w^{t+1}) + 2 \sqrt{\divergence_{\regularizer^t}(\what^{t+1}, \w^{t+1}) \divergence_{\regularizer^t}(\wstar, \what^{t+1})} .$$

Applying Lemmas~\ref{lemma:dualitygap} and~\ref{lemma:maxdivergence}, we get
$$ \divergence_{\regularizer^t}(\wstar, \w^{t+1}) - \divergence_{\regularizer^t}(\wstar, \what^{t+1}) \leq \maxdualitygap + \sqrt{2 \maxdualitygap} (t\problems)^{1/4} \norm{\solutionspace_{max}} .$$

For the third term, using that $\Q^1$ is 1 in exactly one position, we have
$$\insidet{1} \divergence_{\regularizer^1}(\wstar, \w^1) \leq \insidet{1} \frac{1}{2} \norm{\solutionspace_{max}}^2 .$$ 

Combining and dividing by $\eta$, we get the upper bound for the first sum
\begin{align*}
\sum^T_{t=1} \inside \gradient^t \cdot (\what^{t+1} - \wstar) &\leq \frac{1}{\eta} \sum^T_{t=1} \insidet{t+1} \divergence_{\regularizer^{t+1}}(\wstar, \w^{t+1}) - \inside \divergence_{\regularizer^t}(\wstar, \w^{t+1}) \\
&\phantom{\leq} + \frac{1}{\eta} \sum^T_{t=1}  \inside \left( \maxdualitygap + \sqrt{2 \maxdualitygap}(t\problems)^{1/4} \norm{\solutionspace_{max}} \right) \\
&\phantom{\leq} + \frac{1}{2 \eta} \insidet{1} \norm{\solutionspace_{max}}^2.
\end{align*}

\subsection*{Step 2: Second sum}
Similar to step 1, we get
\begin{align*}
\eta \sum^T_{t=1} \outside \gradient^t \cdot (\what^{t+1} - \wstar) &\leq \sum^T_{t=1} \left( \outsidet{t+1} \divergence_{\regularizer^{t+1}}(\wstar, \w^{t+1}) -  \outside \divergence_{\regularizer^t}(\wstar, \w^{t+1}) \right) \\
& \phantom{=} + \sum^T_{t=1} \outside \left( \divergence_{\regularizer^t}(\wstar, \w^{t+1}) - \divergence_{\regularizer^t}(\wstar, \what^{t+1}) \right) \\
& \phantom{=} + \outsidet{1} \divergence_{\regularizer^1}(\wstar, \w^1) .
\end{align*}

As in step 1, we leave the first term unchanged. For the second term, note that $w^{t+1}$ is not project onto $\solutionspace^*$, and thus $\what^t = \w^t$ for all $t$, such that 
$$ \sum^T_{t=1} \outside \left( \divergence_{\regularizer^t}(\wstar, \w^{t+1}) - \divergence_{\regularizer^t}(\wstar, \what^{t+1}) \right) = 0 .$$

For the third term, similar to step 1, we have 
$$\outsidet{1} \divergence_{\regularizer^1}(\wstar, \w^1) \leq \outsidet{1} \frac{1}{2} \norm{\solutionspace_{max}}^2 .$$

Combining, we get the upper bound for the second sum
\begin{align*}
\sum^T_{t=1} \outside \gradient^t \cdot (\what^{t+1} - \wstar) &\leq \frac{1}{\eta} \sum^T_{t=1} \outsidet{t+1} \divergence_{\regularizer^{t+1}}(\wstar, \w^{t+1}) - \outside \divergence_{\regularizer^t}(\wstar, \w^{t+1}) \\
&\phantom{=} + \frac{1}{2 \eta} \outsidet{1} \norm{\solutionspace_{max}}^2.
\end{align*}

\subsection*{Step 3: Combination of steps 1 and 2}
Note that $\inside + \outside = 1$ for all t. Thus, the first terms of step 1 and 2 sum to
$$\frac{1}{\eta} \sum^T_{t=1} \divergence_{\regularizer^{t+1}}(\wstar, \w^{t+1}) - \divergence_{\regularizer^t}(\wstar, \w^{t+1}) .$$

Using Lemma~\ref{lemma:rtplusminusrt}, we get
$$\frac{1}{\eta} \sum^T_{t=1} \divergence_{\regularizer^{t+1}}(\wstar, \w^{t+1}) - \divergence_{\regularizer^t}(\wstar, \w^{t+1}) \leq \frac{1}{2 \eta} \norm{\solutionspace_{max}}^2 \sqrt{T \problems} .$$

Summing the remaining terms and again noting that $\inside + \outside = 1$, we get the upper bound for the first and second sum
\begin{align*}
\sum^T_{t=1} \gradient^t \cdot (\what^{t+1} - \wstar) &\leq \frac{1}{2 \eta} \norm{\solutionspace_{max}}^2 \sqrt{T \problems} + \frac{1}{2 \eta} \norm{\solutionspace_{max}}^2 \\
&\phantom{\leq} + \frac{1}{\eta} \sum^T_{t=1}  \inside \left( \maxdualitygap + \sqrt{2 \maxdualitygap}(t\problems)^{1/4} \norm{\solutionspace_{max}} \right).
\end{align*}

\subsection*{Step 4: Third sum}

To upper bound $\sum^T_{t=1} \good \gradient^t \cdot (\w^t - \what^{t+1})$ we start by using H{\"o}lder's inequality (see for example \cite{beckenbach2012inequalities}) to get
$$\gradient^t \cdot (\w^t - \w^{that+1})  \leq \norm{\gradient^t}^*_{\Q^t} \norm{\w^t - \what^{t+1}}_{\Q^t},$$
where
$$ \norm{\gradient^t}^*_{\Q^t} = \max_{\pmb{x}} \pmb{x} \cdot \gradient^t : \norm{\pmb{x}}_{\Q^t} \leq 1 .$$

For the norm $\norm{\w^t - \what^{t+1}}_{\Q^t}$ we applya Lemma~\ref{lemma:constraint-minimizer} with $\what^{t+1}$ as the constraint minimizer of the objective $\eta \gradient^t \cdot \w + \divergence_{\regularizer^t}(\w, \w^t)$ with $\w^t \in \solutionspace^*$. Using the symmetry of the Bregman divergence for our choice of $\regularizer^t$,
$$ \eta \gradient^t \cdot (\what^{t+1} - \w^t) \leq - 2 \divergence_{\regularizer^t}(\w^{t+1}, \w^t) $$
and thus

$$ \divergence_{\regularizer^t}(\what^{t+1}, \w^t) \leq \frac{1}{2} \eta \gradient^t \cdot (\w^t - \what^{t+1}) .$$

Note that $\divergence_{\regularizer^t}(\w^t, \what^{t+1}) = \frac{1}{2} \norm{\w^t - \what^{t+1}}^2_{\Q^t}$ and thus, 
$$ \norm{\w^t - \what^{t+1}}^2_{\Q^t} \leq \eta \gradient^t \cdot (\w^t - \what^{t+1}) .$$

Using H{\"o}lder's inequality on the right side of the inequality, we get
$$\norm{\w^t - \w^{t+1}}_{\Q^t} \leq \eta \norm{\gradient^t}^*_{\Q^t}.$$
Therefore,
$$ \gradient^t \cdot (\w^t - \w^{t+1}) \leq \eta (\norm{\gradient^t}^*_{\Q^t})^2 $$

We now apply the definition of the dual norm to rewrite $\norm{\gradient^t}^*_{\Q^t}$. Note that $\gradient^t$ is non-zero only in position $\z^t$ and thus
\begin{align*}
\norm{\gradient^t}^*_{\Q^t} &= \max_{\pmb{x}} \pmb{x} \cdot \gradient^t : \norm{\pmb{x}}_{\Q^t} \leq 1 \\
&= \max_{\pmb{x}_{\z^t}} \pmb{x}_{\z^t} \gradient^t_{\z^t} : \left(\norm{\pmb{x}_{\z^t}}^2_2 \sqrt{\counter^t_{\z^t}}\right)^{1/2} \leq 1 \\
&= \max_{\pmb{x}_{\z^t}} \pmb{x}_\z \gradient^t_{\z^t} : \norm{\pmb{x}_{\z^t}}_2 \leq \frac{1}{\sqrt{\counter^t_{\z^t}}^{1/2}}  \\
&\leq \max_{\pmb{x}_{\z^t}} \norm{\pmb{x}_{\z^t}}_2 \norm{\gradient^t_{\z^t}}_2 : \norm{\pmb{x}_{\z^t}}_2 \leq \frac{1}{\sqrt{\counter^t_{\z^t}}^{1/2}}  \\
\end{align*}
The maximum is achieved at $\norm{\pmb{x}_{\z^t}}_2 = \frac{1}{\sqrt{\counter^t_{\z^t}}^{1/2}}$. Thus,
$$ \norm{\gradient^t}^*_{\Q^t} \leq \norm{\gradient^t_{\z^t}}_2 \frac{1}{\sqrt{\counter^t_{\z^t}}^{1/2}} .$$

Inserting, summing, and using Propositions~\ref{proposition:tz} and~\ref{proposition:sumt}, we get the upper bound for the third sum,

\begin{align*}
\sum^T_{t=1}  \good  \gradient^t \cdot (\w^t - \w^{t+1}) &\leq \eta \sum^T_{t=1}  \sum^\problems_{\z=1} \norm{\gradient^t_\z}^2_2 \frac{1}{\sqrt{\counter^t_\z}} \\
&\leq \eta \sum^\problems_{\z=1}  \sum^T_{t=1} \norm{\maxgradient_\z}^2 \frac{1}{\sqrt{\counter^t_\z}}  \\
&\leq 2 \eta \sum^\problems_{\z=1} \norm{\maxgradient_\z}^2  \left( \sqrt{\counter^T_\z} -1 \right) \\
&\leq 2 \eta \norm{\maxgradient_{max}}^2 \sum^\problems_{\z=1} \left( \sqrt{\counter^T_\z} -1 \right) \\
&\leq 2 \eta \norm{\maxgradient_{max}}^2 \sqrt{T\problems} - 2 \eta \norm{\maxgradient_{max}}^2 \problems
\end{align*}

\subsection*{Step 5: Fourth sum}
For the fourth sum we use that 
$$ \gradient^t \cdot (\w^t - \what^{t+1}) = \gradient^t_\zt \cdot (\w^t_\zt - \what^{t+1}_\zt).$$

Using the Cauchy-Schwarz inequality, we get
\begin{align*}
\gradient^t_\zt \cdot (\w^t_\zt - \what^{t+1}_\zt) &\leq \norm{\gradient^t_\zt}_2 \norm{\w^t_\zt - \what^{t+1}_\zt}_2 \\
&\leq \norm{\solutionspace_{max}} \norm{\maxgradient_{max}}
\end{align*}

Thus,
$$\sum^T_{t=1}  \bad \gradient^t \cdot (\w^t - \what^{t+1}) \leq \sum^T_{t=1}  \bad \norm{\solutionspace_{max}} \norm{\maxgradient_{max}}.$$

\subsection*{Step 6: Combination}
Adding the results from steps 1 to 5, we get the result
\begin{align*}
\sum^T_{t=1} \gradient^t \cdot (\w^t - \wstar) &\leq \frac{1}{2 \eta} \norm{\solutionspace_{max}}^2 \sqrt{T \problems} +  2 \eta \norm{\maxgradient_{max}}^2 \sqrt{T\problems} \\
&\phantom{\leq} +\sum^T_{t=1}  \bad \norm{\solutionspace_{max}} \norm{\maxgradient_{max}}\\
&\phantom{\leq} + \frac{1}{\eta} \sum^T_{t=1}  \inside \left( \maxdualitygap + \sqrt{2 \maxdualitygap}(t\problems)^{1/4} \norm{\solutionspace_{max}} \right) \\
&\phantom{\leq} + \frac{1}{2 \eta} \norm{\solutionspace_{max}}^2  - 2 \eta \norm{\maxgradient_{max}}^2 \problems.
\end{align*}
\end{proof}

%!TEX root = ../main.tex

\section{Proof of Corollary ~\ref{corollary:COCP.rare}}\label{appendix_corollaries}
By plugging in specific algorithmic parameters into Theorem~\ref{theorem:COCP} we can get more concrete regret bounds on the \COCP algorithm. We provide no-regret bounds for the parametric choices of sporadic and approximate projection, and note that similar no-regret bounds can also be achieved for different parameters.

\begin{proof}[Proof of Corollary~\ref{corollary:COCP.rare}]

Inserting $\eta = \frac{1}{2} \frac{\norm{\solutionspace_{max}}}{\norm{\maxgradient_{max}}}$ into the results of Theorem~\ref{theorem:COCP}, taking the expected value over $\xi^t$, and using that $\problems \geq 1$, we get
\begin{align*}
\expectedvalue{\sum^T_{t=1} \gradient^t \cdot (\w^t - \wstar)} &\leq 2 \sqrt{T \problems} \norm{\solutionspace_{max}} \norm{\maxgradient_{max}} \\
&\phantom{\leq} + \alpha (1 - \alpha) T \norm{\solutionspace_{max}}  \norm{\maxgradient_{max}} \\
&\phantom{\leq} + 2 \alpha \frac{\norm{\maxgradient_{max}}}{\norm{\solutionspace_{max}}}  \sum^T_{t=1}   \left( \maxdualitygap + \sqrt{2 \maxdualitygap}(t\problems)^{1/4} \norm{\solutionspace_{max}} \right) .
\end{align*}

For the third term, sing $\maxdualitygap = \cbeta (1 - \beta)^2 \frac{\sqrt{\problems}}{\sqrt{t}} \norm{\solutionspace_{max}}^2$ where $\cbeta \geq 0$, $\beta \in [0,1]$, then
\begin{align*}
\maxdualitygap + \sqrt{2 \maxdualitygap}(t\problems)^{1/4} \norm{\solutionspace_{max}} &= \cbeta (1 - \beta)^2 \frac{\sqrt{\problems}}{\sqrt{t}} \norm{\solutionspace_{max}}^2 + (1 - \beta) \sqrt{2 \cbeta} \sqrt{\problems}  \norm{\solutionspace_{max}}^2 \\
&\leq \cbeta (1 - \beta) \sqrt{\problems} \norm{\solutionspace_{max}}^2 + (1 - \beta) \sqrt{2 \cbeta} \sqrt{\problems}   \norm{\solutionspace_{max}}^2 \\
&= (1 - \beta) \sqrt{\problems} (\cbeta + \sqrt{2 \cbeta}) \norm{\solutionspace_{max}}^2 ,
\end{align*}

and, using Proposition \ref{proposition:sumt} for the sum,
$$\sum^T_{t=1} \maxdualitygap +  \sqrt{2 \maxdualitygap}(t\problems)^{1/4} \norm{\solutionspace_{max}}  \leq \alpha (1 - \beta) T \sqrt{\problems} (\cbeta + \sqrt{2 \cbeta}) \norm{\solutionspace_{max}}^2 .$$

Inserting $\alpha = \frac{\calpha}{\sqrt{T}}$  where $\calpha \in [0, \sqrt{T}]$, we get
\begin{align*}
\expectedvalue{\sum^T_{t=1} \gradient^t \cdot (\w^t - \wstar)} &\leq 2 \sqrt{T \problems} \norm{\solutionspace_{max}}  \norm{\maxgradient_{max}} \\
&\phantom{\leq}  + \sqrt{T} \calpha \left(1 - \frac{\calpha}{\sqrt{T}} \right) \norm{\solutionspace_{max}}  \norm{\maxgradient_{max}} \\
&\phantom{\leq} + 2 \sqrt{T \problems} \calpha (\cbeta + \sqrt{2 \cbeta}) (1 - \beta) \norm{\solutionspace_{max}}  \norm{\maxgradient_{max}} .
\end{align*}

\end{proof}
}

%%%%%%%%%%%%%%%%%%%%%%%%%%%%%%%%%%%%%%%%%%%%%%%%%%%%%%%%%
%%%%%%%%%%%%%%%%%%%%%%%%%%%%%%%%%%%%%%%%%%%%%%%%%%%%%%%%% END
\end{document}